\documentclass[10pt,twocolumn,letterpaper]{article}

\usepackage{3dv}
\usepackage{times}
\usepackage{epsfig}
\usepackage{graphicx}
\usepackage{amsmath}
\usepackage{amssymb}

\usepackage{booktabs}
\usepackage{subcaption}
\usepackage{url}

\usepackage{multirow}
\usepackage{colortbl}
\usepackage[dvipsnames]{xcolor}
\usepackage{bbding}
\usepackage[toc,page]{appendix}

\usepackage[pagebackref=true,breaklinks=true,letterpaper=true,colorlinks,bookmarks=false]{hyperref}

\threedvfinalcopy 

\def\threedvPaperID{272} 

\ifthreedvfinal\fi
\begin{document}

\title{PLNet: Plane and Line Priors for Unsupervised Indoor Depth Estimation}

\author{Hualie Jiang$^{1,3}$, Laiyan Ding$^{1,2}$, Junjie Hu$^{2}$, Rui Huang$^{1,2}$\thanks{Corresponding author.}\\
$^{1}$School of Science and Engineering, The Chinese University of Hong Kong, Shenzhen\\
$^{2}$Shenzhen Institute of Artificial Intelligence and Robotics for Society \\
$^{3}$Shenzhen Institute of Advanced Technology, Chinese Academy of Sciences \\
{\tt\small \{hualiejiang, laiyanding\}@link.cuhk.edu.cn, \{hujunjie, ruihuang\}@cuhk.edu.cn}
}

\maketitle
\thispagestyle{empty}

\begin{abstract}
Unsupervised learning of depth from indoor monocular videos is challenging as the artificial environment contains many textureless regions. Fortunately, the indoor scenes are full of specific structures, such as planes and lines, which should help guide unsupervised depth learning. This paper proposes PLNet that leverages the plane and line priors to enhance the depth estimation. We first represent the scene geometry using local planar coefficients and impose the smoothness constraint on the representation. Moreover, we enforce the planar and linear consistency by randomly selecting some sets of points that are probably coplanar or collinear to construct simple and effective consistency losses. 
To verify the proposed method’s effectiveness, we further propose to evaluate the flatness and straightness of the predicted point cloud on the reliable planar and linear regions.
The regularity of these regions indicates quality indoor reconstruction. Experiments on NYU Depth V2 and ScanNet show that PLNet outperforms existing methods. The code is available at \url{https://github.com/HalleyJiang/PLNet}.
\end{abstract}

\vspace{-5pt}
\section{Introduction}

3D reconstruction is a fundamental problem in computer vision, having many applications, such as VR/AR and autonomous robots. Recently, monocular depth estimation (MDE) has been dramatically improved by deep Convolutional Neural Networks (CNNs)~\cite{eigen2014depth, laina2016deeper, fu2018deep, yin2019enforcing} supervised by a large amount of ground truth depth. An alternative is to adopt the photometric consistency of multiple views to guide depth learning. The method that adopts monocular video frames~\cite{zhou2017unsupervised} is attractive, as the monocular videos are prevalent. Such approaches have been actively investigated for the driving scene, such as the KITTI dataset~\cite{geiger2013vision}.

\begin{figure}[t]
\centering{
\input{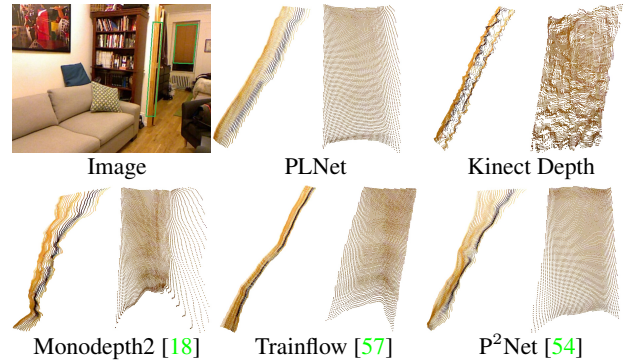}
}
\vspace{-5pt}
\caption{{\bf Visualization of the Flat and Straight Regions.} }
\label{fig:points_vis}
\vspace{-12pt}
\end{figure}

Indoor reconstruction is also of critical importance. But it is challenging to apply the unsupervised framework to the indoor scenario, and only a few attempts have been made. As pointed out by the pioneer work~\cite{zhou2019moving}, indoor videos, such as the NYU Depth V2 dataset~\cite{silberman2012indoor}, have complicated ego-motion, as they are usually recorded by handheld cameras. The problem can be alleviated by sampling the more distant ($\pm10$) frames as the source frames~\cite{zhou2019moving} or weakly rectifying the training sequences~\cite{bian2020unsupervised}. 
Alternatively, we could construct a dataset by moving the camera steadily and sufficiently to solve the problem. The inherent problem is that the indoor scene contains many large textureless regions and some non-Lambertian surfaces, such as the wall, ceiling, and mirror. In these regions, the photometric signal is not discriminative enough to provide supervision.

The man-made scene exhibits strong structural regularity, as it is full of flat and straight elements. 
The related prior knowledge should be utilized in guiding the unsupervised methods but has not been adequately explored. 
Recently, P$^2$Net~\cite{IndoorSfMLearner} extracts homogeneous superpixels~\cite{felzenszwalb2004efficient} and enforces plane regularization. Specifically, P$^2$Net estimates plane parameters of the superpixels by least-square fitting and calculates the difference of the estimated depth and depth of the fitted plane as the loss. However, the homogeneous region is not necessarily planar, \ie, outliers exist. The fitting could be highly biased by outliers and the depth error term of every pixel is problematic. Although robust regression methods exist, such as RANSAC, LMedS, and M-estimator~\cite{meer1991robust}, they iteratively detect or down weight the outliers with manually tuned parameters and it is cumbersome to integrate them into the training of neural networks. 

This paper aims to better infuse the plane and line priors into the unsupervised framework.  First, we represent the scene geometry as local plane coefficients and enforce local smoothness on the coefficients instead of the disparity. The indoor scene is full of planar regions where the coefficients should be constant but not the disparity. 

Second, to better enforce the long-range plane regularization on the segmented superpixels~\cite{felzenszwalb2004efficient}, we propose to establish random sample consistency inspired by the idea of RANSAC (RANdom SAmple Consensus)~\cite{fischler1981random}. 
But we do not iteratively estimate the plane parameters. 
Instead, we just randomly sample some 4-point sets from a homogeneous region and calculate the volume of the parallelepiped expanded from the 4 points as an error term.  In computation, our loss is simpler than that of P$^2$Net~\cite{IndoorSfMLearner} which involves solving equations. More importantly, it is probable that the 4 points are all inliers, and even they contain outliers, it is possible that the consistency error is small. So, our planar consistency loss is potentially more robust (more empirical analysis can be seen in supplementary). 

Third, as the linear segments in the image should usually be straight in 3D, we detect the line segments in images and enforce linear consistency. Following the strategy of the planar consistency, we also randomly select 3 points from a line segment and compute the area of the triangle formed by the 3 points as a consistency error term.

As the indoor scenes are full of planar and linear elements, maintaining the flatness or straightness is essential in reconstruction. Fig.~\ref{fig:points_vis} demonstrates that our PLNet keeps the flatness and straightness much better than existing methods. We also propose to quantitatively evaluate the predicted point cloud's flatness and straightness on the reliable planar and linear regions. We first define the metrics for flatness and straightness inspired by the literature on precision metrology~\cite{gosavi2012form} and point cloud analysis~\cite{hoppe1992surface, fransens2006hierarchical}. Next, we evaluate the Kinect depth on NYU Depth V2~\cite{silberman2012indoor} to find out the reliable planar and linear regions for further evaluation, which also indicates the existence of outliers. 

Our contributions can be summarized as follows. 1) we propose to utilize the plane and line priors to guide the unsupervised indoor depth learning, which includes three technical loss terms, \ie, planar smoothness, planar consistency, and line consistency. 2) we propose to evaluate the depth estimation methods with metrics of flatness and straightness, which reflect the quality of indoor reconstruction. 3) Our PLNet reduces the depth performance gap between the unsupervised method and the supervised counterpart by over $1/3$, and maintains the flatness and straightness much better than existing methods. When generalizing the pre-trained models to ScanNet, PLNet outperforms the supervised one. 

 \section{Related Work}
 
{\textbf{Supervised Depth Estimation} has been extensively studied.} Make3D~\cite{saxena2009make3d} initiates MDE by using the traditional Markov random field trained with image-depth pairs by a 3D laser scanner. However, the accuracy is highly limited until Eigen \textit{et al.}~\cite{eigen2014depth} introduced deep CNNs to this task. Afterward, to improve MDE, many efforts have been made to design better neural models \cite{eigen2015predicting, laina2016deeper, liu2016learning, xu2017multi, hu2021boosting} or introduce better loss function~\cite{fu2018deep, hu2019revisiting, yin2019enforcing}.  Virtual Normal~\cite{yin2019enforcing} is the one close to our work, which enforces the high-order 3D geometric constraints via randomly sampling $N$ groups of 3 points.  We also implement our planar and linear consistency via random sampling. However, the consistency depends on assuming that certain regions are flat or straight instead of the ground truth depth.

\textbf{Unsupervised Depth Estimation} is a more flexible scheme. The supervisory signals are obtained by between-view reconstruction from either stereo images \cite{garg2016unsupervised, godard2017unsupervised} or monocular videos \cite{zhou2017unsupervised} instead of the ground truth depth. This paper mainly refers to the latter, \ie, the unsupervised depth estimation from the more available monocular videos.  
The difficulty of the outdoor scene, such as KITTI~\cite{geiger2013vision}, lies in dynamic objects. Adequate research has been done for this difficulty~\cite{yin2018geonet, bian2019depth, casser2019struct2depth, luo2019every, godard2019digging, klingner2020self, jiang2020dipe, jiang2021unsupervised, lee2021learning}. 
However, only a few works began to apply the unsupervised framework to the more challenging indoor scene~\cite{zhou2019moving, zhao2020towards, bian2020unsupervised, IndoorSfMLearner}. To learn the correspondence well for the textureless regions, Zhou \etal~\cite{zhou2019moving} proposed to predict optical flow in a sparse-to-dense propagation manner and use the optical flow to guide the rigid flow generated by the unsupervised depth learning framework.
To learn the ego-motion better, TrainFlow~\cite{zhao2020towards} estimates the pose by reliable correspondences from a FlowNet with a differentiable two-view triangulation module. 
However, both are limited by optical flow accuracy, and the produced depth maps tend to be over-smooth. 
P$^2$Net~\cite{IndoorSfMLearner} instead adopts patch-based matching on a set of feature points and plane-regularization on precomputed superpixels to achieve better depth maps. However, the regular indoor 3D structures still distort in P$^2$Net's predicted point clouds. Our PLNet recovers the structures much better via enforcing both plane and line priors more reliably. 

\textbf{The Plane and Line Structures} are common indoor elements and are often applied in scene reconstruction or SLAM. The planar structures are usually leveraged in piece-wise reconstruction. The traditional methods~\cite{concha2014using, concha2015dpptam} leverage superpixels~\cite{felzenszwalb2004efficient} to a monocular dense SLAM system, while learning-based methods~\cite{liu2018planenet, yang2018recovering, liu2019planercnn, yu2019single} learn to segment the plane instances and predict the plane parameters supervised by the ground truth segmentation and depth. 
Instead of reconstruction, the line segments are usually utilized to improve the visual odometry or localization in the traditional SLAM system~\cite{zhou2015structslam, pumarola2017pl, gomez2019pl, lee2019elaborate}. However, the important line information has long been overlooked in geometry learning. We only find that VPLNet~\cite{wang2020vplnet} utilizes the detected vanishing points and lines in normal estimation, which indicates that the Manhattan line is useful for scene reconstruction. This paper further shows that even the general line is helpful in recovering the structure regularity.

\section{Method}

\subsection{Basic Model}

This section briefly reviews the basic model of the unsupervised monocular depth estimation~\cite{zhou2017unsupervised, godard2019digging}.
A training sample contains a target view $I_t$ and some temporally adjacent source views $I_s$. There is a DepthNet to estimate the normalized disparity map with the \textit{sigmoid} function.
The disparity will be scaled and shifted to the inverse depth and, finally, the depth map $D$ for the target image. 
Another network, the PoseNet predicts the rigid pose ${\mathbf T}_{t \to s} = \{{\mathbf R}, {\mathbf t}\}$ from the target view $I_t$ to the source view $I_s$. 
Next, with network outputs, the target view can be synthesized from source views. The reconstruction loss is the supervisory signal. 

The synthesis is guided by the two view geometry~\cite{hartley2003multiple}. For a pixel $p_t$ in the target image, to synthesize it, we have to find its corresponding point $p_s$ from the source image,
\begin{equation}
p_s \simeq {\mathbf K}{\mathbf R}D(p_t){\mathbf K}^{-1}p_t + {\mathbf K}{\mathbf t},
\label{eq:rigid}
\end{equation}
where ${\mathbf K}$ is the intrinsic matrix of the moving camera, $\simeq$ is the equality under a scale factor, and the pixel coordinate is expressed in homogeneous coordinates. 
As $p_s$ is usually not on the pixel grid, the synthesis is implemented by the differentiable bilinear sampling operation \cite{jaderberg2015spatial}.
The synthesized result is denoted as $I_{s \to t}$.

The next step is to compute the loss function to train the networks. The primary one is the photometric error between the target view $I_t$ and synthesized view $I_{s \to t}$. Most methods adopt the popular combination of \textit{L}1 and Structural Similarity (SSIM) \cite{wang2004image} proposed by \cite{godard2017unsupervised}, 

\begin{equation}
L_{pe} = \alpha \frac{1 - \mathrm{SSIM}(I_t, I_{s \to t})}{2} + (1-\alpha) \|I_t - I_{s \to t} \|_1,
\label{eq:pe}
\end{equation}
where $\alpha$ is usually set as $0.85$. 

Apart from the photometric error term, to make sure the textureless region can produce plausible depth, an edge-aware disparity smoothness term is also applied in unsupervised training \cite{godard2019digging},  
\begin{equation}
L_{ds} =  \left | \partial_u d^*   \right | e^{-\left | \partial_u I_t \right |} + \left | \partial_v d^*   \right | e^{-\left | \partial_v I_t \right |} , 
\label{eq:dispsmooth}
\end{equation} 
where $d^* = d / \overline{d}$ is the mean-normalized disparity from \cite{wang2018learning} to discourage shrinking of the estimated disparity. 
However, there are also many planar regions in the artificial indoor environment. The disparity or depth in these regions is not constant. To regularize the unsupervised depth learning well, we should mine more reasonable indoor priors.

\subsection{Planar Representation and Smoothness}

We propose to make the DepthNet output the local plane coefficients instead of the conventional normalized disparity. Our planar representation helps perform the smoothness regularization better, as the local plane coefficients should be constant on the planar regions. 

The planar coefficients can be easily converted to depth, not vice versa. Given a pixel coordinate $p$, we can calculate its normalized image coordinate,

\begin{equation}
p^n={\mathbf K}^{-1}p \triangleq (x_n, y_n, 1).
\end{equation}
If the pixel's depth is $Z$, its corresponding 3D point under the camera frame is $(Zx_n, Zy_n, Z)$.
As a camera cannot see the point on a plane that passes through its center, thus we can express an {\bf observable plane} as, 
\begin{equation}
co_xX+co_yY+co_yZ=1,
\label{eq:plane_eq}
\end{equation}
where $co=(co_x, co_y, co_z)$ is the planar coefficient vector. 
By replacing $(X, Y, Z)$ with $(Zx_n, Zy_n, Z)$, we get,
\begin{equation}
1/Z = co_xx_n + co_yy_n + co_z = co^T p^n.
\label{eq:plane2disp}
\end{equation}

From Eqn.~\ref{eq:plane2disp},  we can convert the planar coefficients to the inverse depth, and we can also see that the inverse depth (or disparity) changes with the pixel position in planar regions. 
As the planar coefficients are constant, it is more reasonable to perform smoothness regularization on the planar coefficients than the disparity. The proposed smoothness term on the planar coefficients is computed as,

\begin{equation}
L_{cos} = \sum_{i \in \{x, y, z\}}( \left | \partial_u co_i^*   \right | e^{-\left | \partial_u I_t \right |} + \left | \partial_v co_i^*   \right | e^{-\left | \partial_v I_t \right |} ) , 
\label{eq:cosmooth}
\end{equation} 
where $co_i^* = co_i / \overline{|co_i|}$ is the absolute-mean-normalized coefficient to prevent its shrinking.

\begin{figure}[t]
\begin{center}
\includegraphics[width=0.6\linewidth] {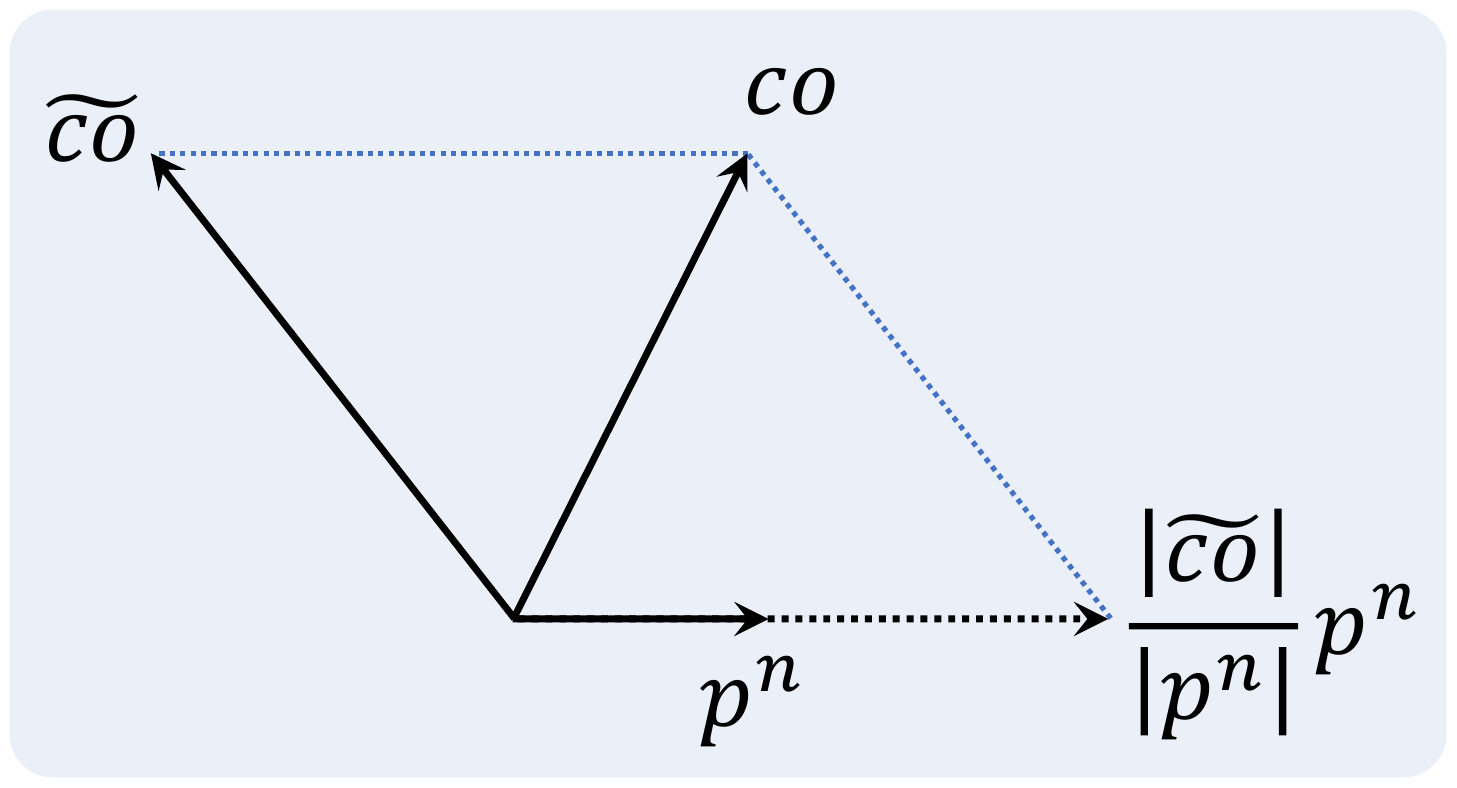}
\end{center}
\vspace{-4mm}
\caption{\textbf{Modulation of the DepthNet Output.}}
\label{fig:coeff_convert}
\end{figure}

However, the DepthNet does not guarantee to produce a positive inverse depth with Eqn.~\ref{eq:plane2disp}, as the output vector may not have an angle less than $90^{\circ}$ with $p^n$. We propose to module the output vector by halving the angle between it (Fig.~\ref{fig:coeff_convert}). Suppose that the output is $\widetilde{co}$, we leverage the normalized image coordinate $p^n$ to convert $\widetilde{co}$ into $co$ by,

\begin{equation}
co = \widetilde{co} + \frac{|\widetilde{co}|}{|p^n|}p^n. 
\label{eq:coeff_convert}
\end{equation}

\subsection{Planar Consistency}
\label{sec:pc}

The indoor scene contains many large textureless regions, such as floor, walls, ceiling, and furniture. 
The photometric values of these regions are not discriminative enough, thus the unsupervised depth learning cannot be well guided by the photometric error. 
Besides the local planar smoothness, we also establish long-range planar consistency on piece-wise non-texture regions.

\begin{figure}[t]
\centering{
\input{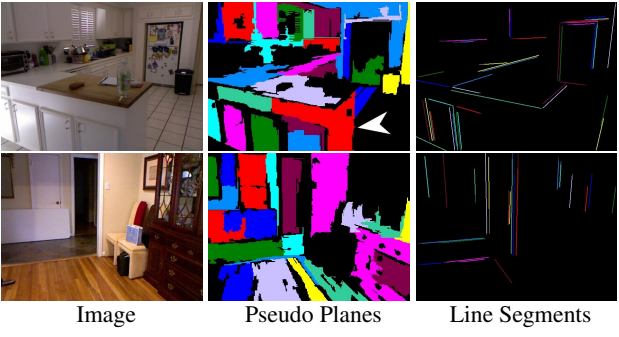}
}
\caption{{\bf Examples of Pseudo Planes and Line Segments.} }
\label{fig:plane_line_example}
\end{figure}

We follow P$^2$Net~\cite{IndoorSfMLearner} to extract the homogeneous regions in images using the Felzenszwalb superpixel segmentation algorithm~\cite{felzenszwalb2004efficient}, and keep regions larger than 1000 pixels. 
Fig.~\ref{fig:plane_line_example} demonstrates some examples of segmented regions, where we can see that most of them are planes. However, outliers exist. For example, the two perpendicular sides of the cabinet in the first row of Fig.~\ref{fig:plane_line_example} are segmented as the same region (the red region pointed by a white arrow). 
Therefore, we call these regions the pseudo plane. 

\begin{figure}[t]
\begin{center}
\includegraphics[width=0.8\linewidth] {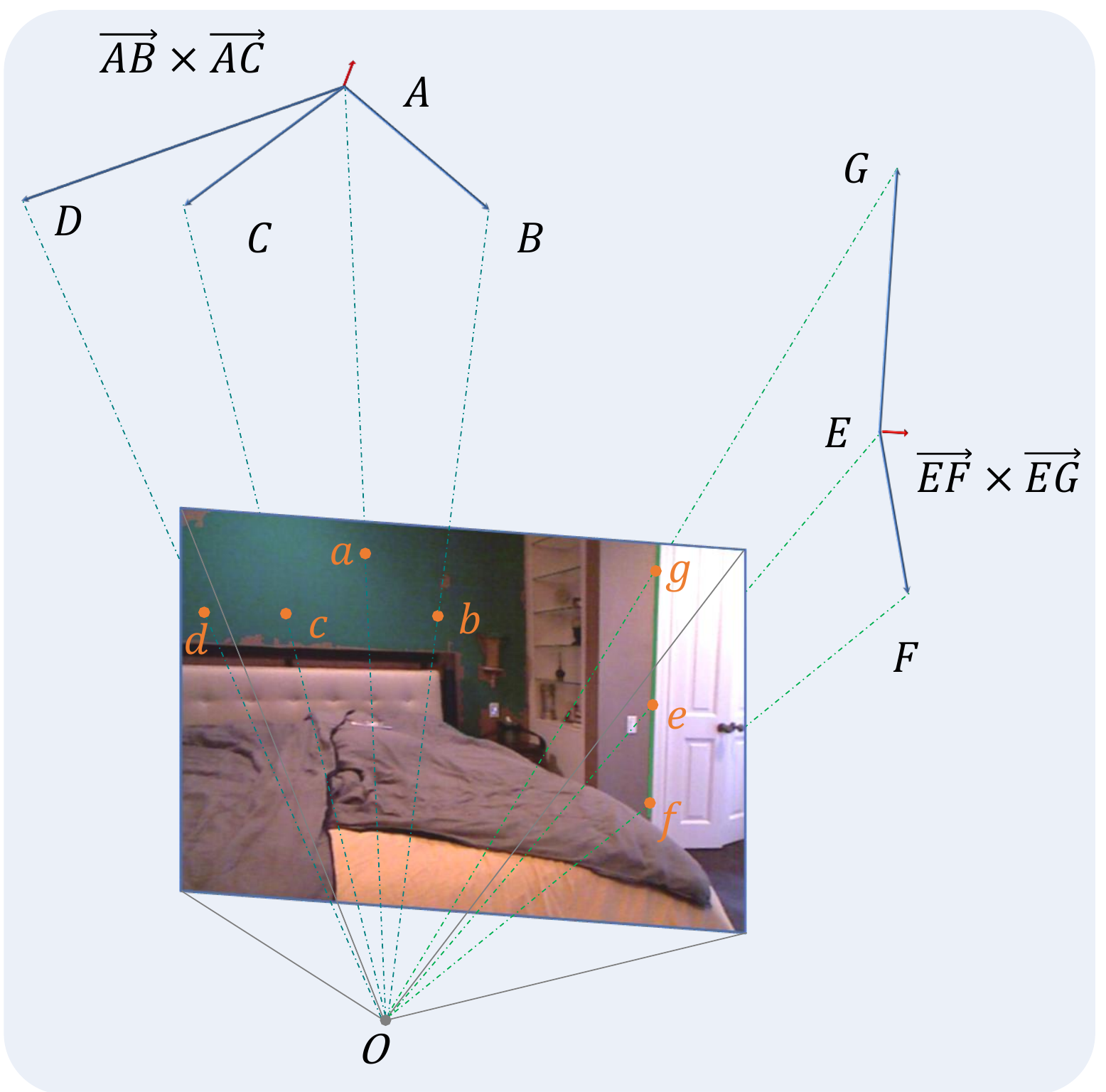}
\end{center}
\vspace*{-4mm}
\caption{\textbf{Planar and Linear Consistency.} }
\label{fig:scatter}
\end{figure}

We thus propose a potentially more robust plane consistency loss. Our scheme is illustrated in Fig.~\ref{fig:scatter}. For a pseudo plane, we randomly sample 4 pixels, \ie, $a$, $b$, $c$ and $d$. With the estimated depth, we can reproject the pixels to 3D points, $A$, $B$, $C$ and $D$. The cross product of
$\overrightarrow{AB}$ and $\overrightarrow{AC}$ will be perpendicular to the plane of $\bigtriangleup ABC$. Additionally, $D$ is expected to be co-planar to $\bigtriangleup ABC$ and the dot product of $\overrightarrow{AD}$ and $\overrightarrow{AB} \times \overrightarrow{AC}$ should be $0$. Therefore, $\left| \overrightarrow{AB} \times \overrightarrow{AC} \cdot \overrightarrow{AD} \right|$ can work as an error term. 
Specifically, given an input target image, we randomly select $N_p$ 4-point sets from the pseudo-planes in total. The number of sets assigned to a pseudo-plane is proportional to its number of pixels. The proposed plane consistency loss is computed by, 

\begin{equation}
L_{pc} = \frac{1}{N_p}\sum_{i =1}^{N_p} \left| \overrightarrow{A_iB_i} \times \overrightarrow{A_iC_i} \cdot \overrightarrow{A_iD_i} \right|.
\label{eq:plane_cons}
\end{equation} 

Compared with the scheme of P$^2$Net~\cite{IndoorSfMLearner} which fits the pseudo planes with the least square method by solving equations, our approach is simpler in computation.

\subsection{Linear Consistency}
\label{sec:lc}
The line structure is another pervasive element in indoor scenes. 
The only case when a 3D curve projects to a 2D line segment is that the 3D curve lies on a plane, and the plane passes through the camera center. 
Therefore, the line segment in the 2D image always corresponds to the straight line in 3D. 
However, we observe that the state-of-the-art unsupervised depth estimation methods produce distorted results for the lines. 
Thus, we propose to utilize this prior knowledge to regularize unsupervised depth learning. 

Extracting line segments from images is well studied in computer vision. We adopt the Line Segment Detector (LSD)~\cite{von2008lsd, von2012lsd} to extract the line segments. We abandon line segments shorter than $1/10$ of the image diagonal, as short line segments are not very valuable for consistency. The outputs of LSD are the pair endpoints of line segments; thus, we have to judge which pixel belongs to a specific line segment. 
We assign a pixel to a line segment if the pixel's distance to the line segment is smaller than 1 unit. Fig.~\ref{fig:plane_line_example} shows some examples of extracted line segments.

\begin{figure*}[t]
\begin{center}
\includegraphics[width=\linewidth] {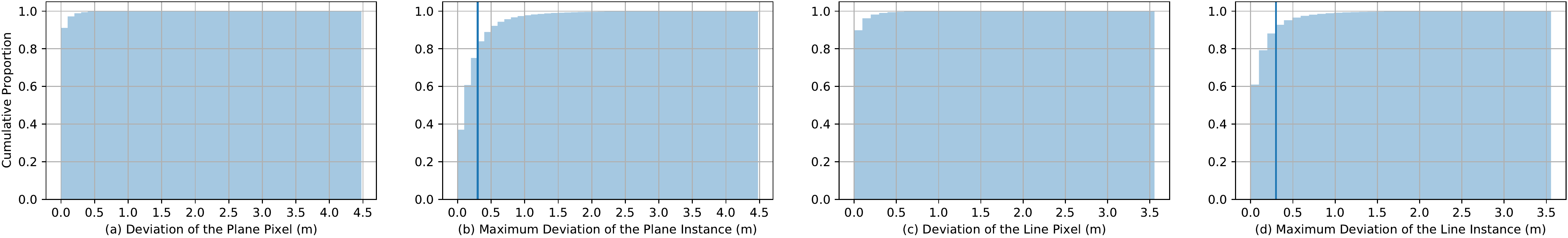}
\end{center}
\vspace*{-5mm}
\caption{\textbf{Distribution of Kinect Depth's Plane and Line Deviations.}}
\label{fig:deviation}
\vspace*{-2pt}
\end{figure*}

The linear consistency loss still follows the strategy of the planar consistency loss. Namely, we randomly sample 3 pixel from a line segment, \ie, $e$, $f$ and $g$. Their corresponding 3D points, $E$, $F$, and $G$, should be constrained on the same line. The cross product of $\overrightarrow{EF}$ and $\overrightarrow{EG}$ should be a zero-vector. Thus,  $\left|\overrightarrow{EF} \times \overrightarrow{EG}\right|$ can be an error term. Likewise, we randomly select $N_l$ 3-point sets from the line segments of the target image. The number of sets assigned to a line segment is proportional to its number of pixels.  The proposed linear consistency loss is computed as,

\begin{equation}
L_{lc} = \frac{1}{N_l}\sum_{i =1}^{N_l}  \left|\overrightarrow{E_iF_i} \times \overrightarrow{E_iG_i}\right|. 
\label{eq:line_cons}
\end{equation} 

\subsection{Overall Loss Function}
The integrated objective function is combined with the photometric error loss $L_{pe}$, the plane smoothness loss $L_{cos}$, the plane consistency loss $L_{pc}$ and the line consistency loss $L_{lc}$, which is computed as,

\begin{equation}
L = L_{pe} + \alpha_{cos} L_{cos} + \alpha_{pc} L_{pc} + \alpha_{lc} L_{lc}, 
\label{eq:overall_loss}
\end{equation} 
where $\alpha_{cos}$, $\alpha_{pc}$ and $\alpha_{lc}$ are weights for the regularization losses, which are set to 0.2, 2.0 and 0.5, respectively. $N_p$ in $L_{pc}$ and $N_l$ in $L_{lc}$ are set to $512$ and $128$, respectively. The overall framework is illustrated in the supplementary.

\section{Flatness and Straightness}
\label{sec:flat_and_strg}

As the indoor environment contains many planes and lines, the flatness and straightness of the estimated point cloud on these regions should be an essential requirement for high-quality reconstruction. As our approach is based on the plane and line priors, evaluating the flatness and straightness can verify our approach's effectiveness. Therefore, we propose to perform such an evaluation on a reliable part of the pseudo planes and line segments.

\subsection{Metrics}

The deviation is the metric measuring the form tolerance in precision metrology~\cite{gosavi2012form}. The deviation is the distance from the measured point to the fitted form (plane or line). We can use the \textit{average deviation} (Avg Dev) of all points and the \textit{maximum deviation} (Max Dev) of a form instance to evaluate the tolerance.
As the deviation is scale-variant and the unsupervised monocular depth estimation cannot recover the actual scale, we first align the predicted depth to the ground-truth depth following the conventional practice~\cite{zhou2017unsupervised}. In fitting the point cloud to a plane or a line, we adopt the total least squares method~\cite{nievergelt1994total}.

Besides the deviation metrics, we propose to use Principal Component Analysis (PCA)~\cite{abdi2010principal} to evaluate the distribution of the estimated point cloud.
PCA is a conventional tool in point cloud analysis, such as calculating normals~\cite{hoppe1992surface} and extracting planar patches~\cite{fransens2006hierarchical}. The PCA of a point cloud yields 3 principal orthonormal directions, $v_1$, $v_2$ and $v_3$, and their associate principal values, $\lambda_1$, $\lambda_2$ and $\lambda_3$ ($\lambda_1 \geq \lambda_2 \geq \lambda_3$). The principal value represents the variance when projecting the data onto its direction, and the former principal direction accounts for the most variance as possible. Therefore, to evaluate the flatness, we can use the following formula as the residual ratio when projecting a 3D points cloud into the principal 2D plane,
\begin{equation}
R_{plane} = \lambda_3 / (\lambda_1 + \lambda_2 + \lambda_3). 
\label{eq:plane_res_ro}
\end{equation} 
Similarly, to evaluate the straightness, we use the residual ratio when projecting 3D points into the principal 1D line,
\begin{equation}
R_{line} = (\lambda_2 + \lambda_3) / (\lambda_1 + \lambda_2 + \lambda_3). 
\label{eq:line_res_ro}
\end{equation} 

In summaries, while the deviations represent the pixel-wise error, the residual ratios derived from PCA describe how the point cloud like the plane or line. 

\begin{figure*}[t]
\centering{
\input{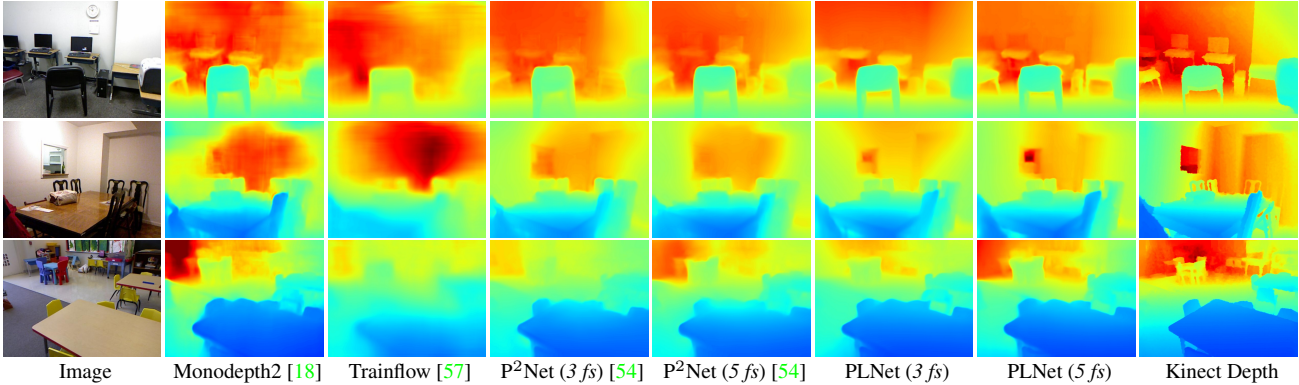}
}
\vspace{-2.5pt}
\caption{{\bf Qualitative Results of Predicted Depth Maps.} Best viewed when zooming in.}
\label{fig:depths}
\vspace{2.5pt}
\end{figure*}

\begin{table*}[t]
  \centering
  \resizebox{0.6\textwidth}{!}{
  \begin{tabular}{l| c | c c c  | c c c}
     \toprule[1pt]
     \multirow{2}{*}{Methods} & \multirow{2}{*}{Supervised} & \multicolumn{3}{c|}{Error Metrics $\downarrow	$} & \multicolumn{3}{c}{Accuracy metrics $\uparrow	$}  \\
     \cline{3-8}
      & & rel & log10 & rms  & $\delta < 1.25$ & $\delta < 1.25^2$ & $\delta < 1.25^3 $ \\
     \hline
     Make3D~\cite{saxena2009make3d} & \Checkmark  &  0.349 & - & 1.214 & 0.447 & 0.745 & 0.897 \\
     
     Eigen~\etal~\cite{eigen2014depth} & \Checkmark  & 0.215 & - & 0.907  & 0.611 & 0.887 & 0.971\\
     
     Liu~\etal~\cite{liu2016learning} & \Checkmark  & 0.213 & 0.087 & 0.759 & 0.650 & 0.906 & 0.976 \\
     Eigen~\etal~\cite{eigen2015predicting} & \Checkmark  & 0.158 & - & 0.641 & 0.769 & 0.950 & 0.988 \\
     FCRN~\cite{laina2016deeper} & \Checkmark  & 0.127 & 0.055 & 0.573 & 0.811 & 0.953 & 0.988 \\
     Xu ~\etal~\cite{xu2017multi} & \Checkmark  & 0.121 & 0.052 & 0.586 & 0.811 & 0.954 & 0.987 \\
     DORN~\cite{fu2018deep} & \Checkmark  & {0.115} & 0.051 & {0.509} & 0.828 & 0.965 &  0.992 \\
     Hu~\etal~\cite{hu2019revisiting} & \Checkmark  & {0.115} & {0.050} & 0.530 & {0.866} & {0.975} & {0.993} \\
     Yin~\etal~\cite{yin2019enforcing}  & \Checkmark  & {0.108} & {0.048} & {0.416} & {0.875} & {0.976} & {0.994} \\
     \hline

     MovingIndoor~\cite{zhou2019moving} & \XSolidBrush & 0.208 & 0.086 & 0.712 & 0.674 & 0.900 & 0.968 \\
     TrainFlow~\cite{zhao2020towards} & \XSolidBrush & 0.189 & 0.079 & 0.686 & 0.701 & 0.912 & 0.978 \\
     MonoDepth2~\cite{godard2019digging}& \XSolidBrush & 0.171 & 0.071 & 0.639 & 0.750 & 0.945 & 0.987  \\

     P$^2$Net~\cite{IndoorSfMLearner} (\textit{3 frames}) & \XSolidBrush &  {0.159} & {0.068} & {0.599} & {0.772} & {0.942} & {0.984} \\
      P$^2$Net~\cite{IndoorSfMLearner} (\textit{5 frames}) & \XSolidBrush &  {0.150} & {0.064} & {0.561} & {0.796} & {0.948} & {0.986} \\
     \hline
     PLNet (\textit{3 frames}) & \XSolidBrush & {0.151} & {0.064} & {0.562} & {0.790} & {0.953} & {0.989} \\
     PLNet (\textit{5 frames}) & \XSolidBrush &  \textbf{0.144} & \textbf{0.061} & \textbf{0.540} & \textbf{0.807} & \textbf{0.957} & \textbf{0.990} \\
     \hline
     DepthNet & \Checkmark &   0.136  &  0.058  &   0.492   &   0.828  &   0.963  &   0.990  \\ 
     \toprule[1pt]
    \end{tabular}
  }
  \hfill
  \raisebox{0pt}{
  \begin{minipage}[c]{0.38\textwidth}
  \caption{{\bf Performance Comparison on the NYU Depth V2~\cite{silberman2012indoor} Dataset.} The first block shows the results of supervised methods, and the second block shows the results of unsupervised methods. The third block is the results of our approach. \textit{3 frames} indicate that 2 adjacent frames are used as source views for training, while \textit{5 frames} indicate 4 adjacent frames are used. We train a supervised DepthNet with the BerHu~\cite{laina2016deeper} in the last block. As the DepthNet backbone is shared among Monodepth2~\cite{godard2019digging}, TrainFlow~\cite{zhao2020towards}, P$^2$Net~\cite{IndoorSfMLearner} and PLNet, the supervised DepthNet can be regarded as an upper bound of these unsupervised models trained with monocular videos. }
  \label{tab:nyudepth}
  \end{minipage}}
\end{table*}

\subsection{Evaluation on Kinect Depth}

To evaluate flatness and straightness on the predicted point cloud, we should find which regions belong to the plane or line first. As manually pixel-wise labeling is time-consuming, therefore we evaluate the flatness and straightness of the ground-truth depth on the detected pseudo-planes and line segments and select the most reliable instances for assessing the depth estimation methods. Besides, the evaluation of the Kinect depth provides quantitative verification of the assumption of our approach.

We evaluate the popular dataset for indoor depth estimation, NYU Depth V2~\cite{silberman2012indoor}. 
As the whole dataset is quite large, we only use the labeled subset (795 for training, 654 for the test). We use the methods mentioned in Sec.~\ref{sec:pc} and Sec.~\ref{sec:lc} to extract the pseudo-planes and line segments. The average numbers of extracted planes and lines per image are 18.8 and 25.8.  We evaluate the subset with the deviation metrics, and the distributions of deviations are illustrated in Fig.~\ref{fig:deviation}. (a) and (c) are the distributions of the pixels' deviations of all the pseudo planes and line segments. (b) and (d) are the distributions of the maximum deviations of the instances of pseudo-planes and line segments.

Most deviations are smaller than 0.1m, and the average deviations of planes and lines are 0.036m and 0.040m. The deviations mainly come from the inherent noise of the Kinect camera~\cite{zhang2012microsoft}, which is in centimeters~\cite{khoshelham2011accuracy}. However, the most significant deviation is about 4 meters, indicating that the extracted planes and lines contain outliers. According to our observation, the outliers of the pseudo-planes come from incorrect superpixel segmentation. 
Some line segments lie in the depth discontinuity edges, where depth and RGB cannot be perfectly aligned. Thus, these line segments may present high deviations with Kinect depth. 

To better evaluate the predicted point cloud, we have to select the reliable pseudo planes and line segments. We only retain the instance whose maximum deviation smaller than 0.3m. As shown in (b) and (d) of Fig.~\ref{fig:deviation}, about $75\%$ of pseudo planes and $88\%$ are selected for evaluating the depth estimation methods. After the selection, the average deviations of planes and lines become 0.024m and 0.020m.

\section{Experiments}

\subsection{Experimental Details.} 

We perform experiments on the popular NYU Depth V2 dataset~\cite{silberman2012indoor} for indoor depth estimation. We sample the raw video by 5 frames as the target view and obtain a training set of about 47K from the training scenes. We follow previous works~\cite{zhou2019moving, IndoorSfMLearner} to sample the video by $\pm10$ frames as source frames to obtain enough translation. We use the official 654 test samples in the evaluation. In evaluation, apart from the common metrics for depth estimation~\cite{liu2016learning,laina2016deeper}, we also evaluate the unsupervised methods with the proposed flatness and straightness metrics in Sec.~\ref{sec:flat_and_strg}.  As unsupervised methods cannot recover the absolute scale, we align the prediction with the ground truth following SfMLearner~\cite{zhou2017unsupervised}.

We use the single-scale version of the baseline of Monodepth2~\cite{godard2019digging} as the basic model in this paper. We empirically find Monodepth2's three techniques for the outdoor scene to not work for the indoor scene. There is no noticeable performance difference between the single-scale and multi-scale baselines. We also train a supervised DepthNet for comparison. We implement the models using Pytorch on an Nvidia TITAN Xp GPU. The ResNet18~\cite{he2016deep} pre-trained on ImageNet~\cite{deng2009imagenet} is used as the backbone for both DepthNet and PoseNet. The input size of networks is $288\times384$, and we up-sample the predicted depth to the original resolution before evaluation. We train the networks for 20 epochs with a batch size of 12 with the Adam~\cite{kingma2014adam} optimizer. The learning rate is $10^{-4}$ initially and divided by 10 after 15 epochs.  The training takes about 8 hours.

\begin{figure*}[t]
\centering{
\input{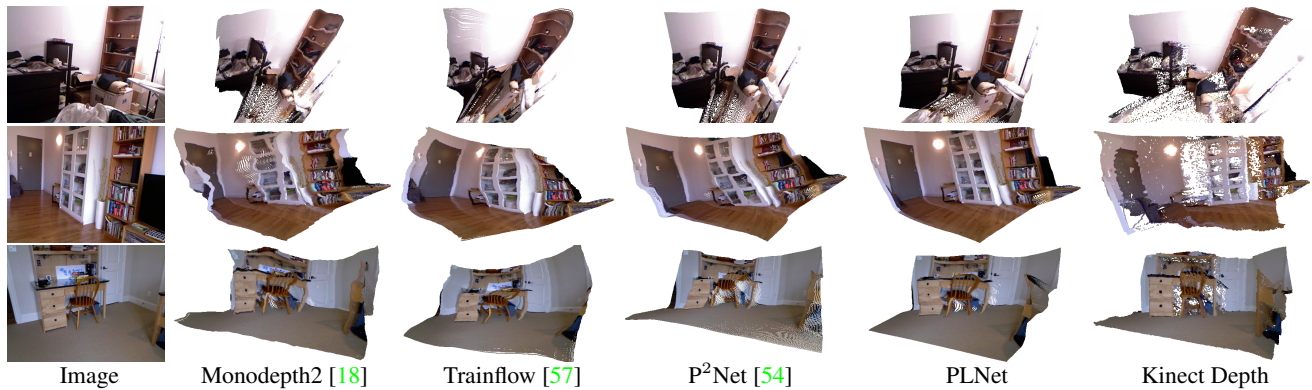}
}
\caption{{\bf Qualitative Results of Predicted Point Clouds.} Best viewed when zooming in.}
\label{fig:points}
\vspace*{5pt}
\end{figure*}

\begin{table*}[t]
  \centering
  \resizebox{0.60\textwidth}{!}{
  \begin{tabular}{l| c c c  | c c c}
     \toprule[1pt]
\multirow{2}{*}{Methods}  & \multicolumn{3}{c|}{Flatness Metrics $\downarrow$ } & \multicolumn{3}{c}{Straightness Metrics $\downarrow$ }  \\
\cline{2-7}
& Avg Dev & Max Dev &  $R_{plane}$  &  Avg Dev & Max Dev &  $R_{line}$ \\
     \hline
     MonoDepth2~\cite{godard2019digging} & 0.0387 & 0.1479 & 0.0248 & 0.0344 & 0.1124 & 0.0428 \\
     TrainFlow~\cite{zhao2020towards} & 0.0290 & \textbf{0.1007} & 0.0146 & 0.0256 & 0.0757 & 0.0221  \\
     P$^2$Net~\cite{IndoorSfMLearner} (\textit{3 frames})  & 0.0191 & {0.1010} & 0.0127 &  0.0165 & 0.0586 & 0.0170 \\
      P$^2$Net~\cite{IndoorSfMLearner} (\textit{5 frames})  & 0.0224 &  0.1063 &  0.0148 &  0.0172& 0.0602 & 0.0171\\
     \hline
     PLNet (\textit{3 frames}) &  \textbf{0.0163} & {0.1027} & \textbf{0.0123} & \textbf{0.0117}  & \textbf{0.0459} & \textbf{0.0118} \\
     PLNet (\textit{5 frames})  & 0.0198 & 0.1154 & 0.0153 & 0.0147 & 0.0591 & 0.0158 \\
     \hline
     Supervised DepthNet  &  0.0252    &  0.1362    &  0.0194    &  0.0257    &  0.0863    &  0.0367    \\
     Kinect Depth  &   0.0194  &  0.1170  &   0.0173   &  0.0234  &    0.0808  &   0.0373  \\ 
     \toprule[1pt]
    \end{tabular}
  }
  \hfill
  \raisebox{0pt}{
  \begin{minipage}[c]{0.38\textwidth}
  \caption{ {\bf Quantitative Evaluation of Flatness and Straightness on the NYU Depth V2 ~\cite{silberman2012indoor} Dataset. } The first block reports the results of other unsupervised methods. The second block presents Our PLNet's results. The last block reports the supervised DepthNet and the ground truth depth obtained by the Kinect camera. }
  \label{tab:nyu_flat_stright}
  \end{minipage}}
\end{table*}

\subsection{Depth Evaluation}

The depth quantitative evaluation results using Kinect depth as ground truth are reported in Tab.~\ref{tab:nyudepth}.  Our PLNet remarkably outperforms the latest unsupervised method, P$^2$Net~\cite{IndoorSfMLearner}. Specifically, PLNet reduces the performance gap between P$^2$Net~\cite{IndoorSfMLearner} to the supervised DepthNet by over $1/3$. The qualitative results of the unsupervised methods are also demonstrated in Fig.~\ref{fig:depths}. Monodepth2~\cite{godard2019digging} produces depth maps with many artifacts, while Trainflow~\cite{zhao2020towards} tends to over smooth the results. P$^2$Net~\cite{IndoorSfMLearner} predicts cleaner depth maps, but PLNet further produces more accurate depth and clear object boundaries. 

\subsection{Point Cloud Evaluation}

As mentioned before, the indoor scene contains adequate planar and linear regions, and it is crucial to maintain the flatness and straightness for these regions in 3D reconstruction. Therefore, to assess the quality of the produced point cloud by the depth estimation methods, we perform the evaluation described in Sec.~\ref{sec:flat_and_strg}. 

The quantitative results are listed in Tab.~\ref{tab:nyu_flat_stright}. As the Kinect depth contains certain noise, its corresponding point cloud's flatness and straightness are not very good. The supervised DepthNet does even worse as it is guided by the noisy Kinect depth. The unsupervised Monodepth2~\cite{godard2019digging}, which mainly depends on the photometric error, significantly underperforms the supervised DepthNet. Another unsupervised method, Trainflow~\cite{zhao2020towards}, whose DepthNet is guided by the sparse triangulated depth by the unsupervised optical flow estimation, has a comparable performance with the supervised DepthNet. Trainflow also achieves the lowest Max Dev for flatness, as its depth maps are quite smooth, and no sharp point cloud is produced.

P$^2$Net~\cite{IndoorSfMLearner} achieves better results than the two previous unsupervised methods, as it enforces piece-wise planar consistency. Both metrics of flatness and straightness are reduced; thus, we hypothesize that the planar consistency is also helpful to straightness, which can also be verified by the experiments in Sec.~\ref{sec:abla}. 
With the adoption of both adjacent planar smoothness and long-range planar consistency, our PLNet outperforms P$^2$Net on the flatness. Furthermore, as we introduce the linear consistency, our PLNet surpasses P$^2$Net on the straightness to a larger extent. As the farther source frames have bigger translations with the target view, PLNet and P$^2$Net can achieve better depth estimation results when trained with 5 frames according to triangulation, which is also verified by Tab.~\ref{tab:nyudepth}. 
However, adding more frames does not necessarily indicates better performance on flatness and straightness, as more photometric losses could cause difficulties in plane and line regularization.

Fig.~\ref{fig:points} shows the predicted point clouds, which are rendered with Open3D~\cite{zhou2018open3d}. The results of PLNet and P$^2$Net are trained with \textit{3 frames}, for which they are better on flatness and straightness. PLNet recovers the structures better than other methods, especially for the line structures.

\begin{table*}[t!]
  \centering
  \resizebox{\textwidth}{!}
{
    \footnotesize
    \begin{tabular}{l|c|l|c c c|c c|c c}
  \toprule
 \multirow{2}{*}{DepthNet Output} &  \multirow{2}{*}{ID} & \multirow{2}{*}{Methods}
 & \multicolumn{3}{c|}{Depth Metrics} &     \multicolumn{2}{c|}{Flatness Metrics} & \multicolumn{2}{c}{Straightness Metrics}
 \\
 \cline{4-6}  \cline{7-8}   \cline{9-10}
 & & &   rel $\downarrow$ & rms $\downarrow$ & $\delta < 1.25$ $\uparrow$ & Avg Dev  $\downarrow$ & $R_{plane}$  $\downarrow$ & Avg Dev $\downarrow$& $R_{line}$  $\downarrow$ \\
  \hline
  \multirow{4}{*}{Normalized Disparity} 
  & (a) & $L_{pe}$ & 0.163 & 0.610 & 0.764  & 0.0353 & 0.0232 & 0.0281 & 0.0349 \\
  & (b) & $L_{pe}+L_{ds}$ & 0.161 & 0.601 & 0.769 & 0.0343 & 0.0231 & 0.0282 & 0.0354 \\
  & (c) & $L_{pe}+L_{ds}+L_{pc}$ & 0.157 & 0.579 & 0.775 & 0.0194 & 0.0153 & 0.0202 & 0.0229 \\
  & (d) & $L_{pe}+L_{ds}+L_{pc}+L_{lc}$ & 0.154 & 0.565 & 0.784  & 0.0174 & 0.0146 & 0.0120 & 0.0126 \\
\hline
\multirow{8}{*}{Planar Coefficients} 
& (e) &  $L_{pe}$& 0.161 & 0.603 & 0.773 & 0.0368 & 0.0258 & 0.0322 & 0.0446  \\

& (f) & $L_{pe}+L_{cos}$ &   0.157 & 0.587 & 0.778  &  0.0297 &  0.0174  &  0.0241 &  0.0263    \\
& (g) & $L_{pe}+L_{pc}$ &   0.155 & 0.575 &   0.781  &  0.0196  &  0.0168  & 0.0218 &  0.0274    \\
& (h) & $L_{pe}+L_{cos}+L_{pc}$ &   0.154  &  0.573  &  0.785  &  0.0176  &  0.0132  &  0.0184   &  0.0197  \\
& (i) & $L_{pe}+L_{lc}$ &   0.164 &  0.645 &   0.768  &  0.0384  &  0.0215  &  0.0157  &  0.0163  \\
& (j) & $L_{pe}+L_{cos}+L_{lc}$ &   0.154  & 0.575  &  0.786  &  0.0272  & 0.0160  &  0.0125 &  0.0119  \\
& (k) & $L_{pe}+L_{pc}+L_{lc}$ &   0.153 &   0.573  &  0.787 &  0.0169  &  0.0145  &  0.0119 &  0.0126  \\
& (l) & $L_{pe}+L_{cos}+L_{pc}+L_{lc}$ & \textbf{0.151} & \textbf{0.562} & \textbf{0.790} & \textbf{0.0162} & \textbf{0.0123} & \textbf{0.0117} & \textbf{0.0118} \\
\bottomrule
\end{tabular}
  }
\caption{\textbf{Ablation Study.} The two blocks report the results of models with two different types of DepthNet output. }
\label{tab:ablation}
\end{table*}

\subsection{Ablation Study}
\label{sec:abla}

We perform ablation experiments to examine the effectiveness of our technical designs, and the results are listed in Tab.~\ref{tab:ablation}. All models are trained with \textit{3 frames}. The weight for $L_{ds}$ is conventionally set to 0.001~\cite{godard2019digging, IndoorSfMLearner}. Model (b) is the single-scale version of the baseline of Monodepth2~\cite{godard2019digging}, which performs even better than Monodepth2. After enforcing the proposed planar consistency $L_{pc}$ and linear consistency $L_{lc}$ upon (b), the depth estimation performance is markedly enhanced.  After adopting the proposed planar representation and smoothness, \ie, the model (m), the performance is further improved. Therefore, the proposed techniques improve the unsupervised depth estimation. 

When examining the effectiveness of our techniques for flatness and straightness, we can find that our methods significantly reduce the average deviation and residual ratio for both flatness and straightness. As expected, the planar and linear consistency improve substantially the flatness and straightness, respectively. In particular, the planar and linear consistency halve their respective average deviation and residual ratio when added to the base model (f). The planar smoothness can also improve the flatness significantly. The planar smoothness and consistency also have some positive effects on the straightness. It is the same for the linear consistency on the flatness. 

\subsection{Generalization}

To examine our approach's generalization ability, we follow P$^2$Net to transfer the pre-trained models on NYU Depth V2 to another large-scale indoor RGBD dataset, ScanNet~\cite{dai2017scannet}. We evaluate the methods on the 553 testing images randomly picked from diverse scenes of ScanNet, which is provided by P$^2$Net. The results are listed in Tab.~\ref{tab:transfer}. PLNet still performs better than other unsupervised methods.

\begin{table}[ht] 
\centering
\resizebox{0.88\columnwidth}{!}
{
\begin{tabular}{l|ccc|c}
  \toprule
  Methods   &  rel $\downarrow$ & rms $\downarrow$ &  log10 $\downarrow$ & $ \delta < 1.25$ $\uparrow$ \\ \hline
  MovingIndoor~\cite{zhou2019moving}  & 0.212 & 0.483 & 0.088 & 0.650  \\
  Trainflow~\cite{zhao2020towards} & 0.195 & 0.458 & 0.083 & 0.684 \\
  Monodepth2~\cite{godard2019digging} & 0.191 & 0.444  & 0.080 & 0.692  \\
  P$^2$Net (\textit{3 frames})~\cite{IndoorSfMLearner} & 0.178 & 0.425 & 0.075 & 0.729  \\
  P$^2$Net (\textit{5 frames})~\cite{IndoorSfMLearner} & {0.175} & {0.420} & {0.074} & {0.740}  \\
  \hline
  PLNet (\textit{3 frames}) & 0.176 & 0.414 & 0.074 & 0.735  \\
  PLNet (\textit{5 frames}) & \textbf{0.168} & \textbf{0.404} & \textbf{0.072} & \textbf{0.750}  \\
  \hline
  Supervised DepthNet & {0.171} & {0.420} & {0.073} & {0.740}  \\
  \bottomrule
\end{tabular}
}
\caption{\textbf{Performance Comparison on Generalization.} Models are trained on NYUv2 and evaluated on ScanNet.}
\label{tab:transfer}
\end{table}

We also include the results of the supervised DepthNet. We perform the prediction and ground truth alignment~\cite{zhou2017unsupervised} for the supervised DepthNet before evaluation, following the unsupervised methods. The reason is that when disabling the alignment, the supervised DepthNet performs much worse to the unsupervised methods. We find that the scale of the predicted depth by the supervised DepthNet is far from the ground-truth scale for some test images. Nevertheless, we are surprised to see that our PLNet trained with \textit{5 frames} markedly outperforms the supervised DpethNet. 

\section{Conclusion}
In this paper, we propose a simple and effective method that applies the prior knowledge of the plane and line to enhance unsupervised indoor depth estimation. We have technically developed three loss functions to guide the unsupervised learning, including a planar smoothness loss, a planar consistency loss, and a linear consistency loss. Being aware of the importance of flatness and straightness in indoor reconstruction, we also propose to evaluate the flatness and straightness of the estimated results, apart from the conventional depth estimation evaluation. The experimental results verify the effectiveness of the proposed approach. In the future, we would explore in more priors, such as parallel and orthogonal relationships of the line and plane, into unsupervised indoor depth learning.

{\small
\bibliographystyle{ieee_fullname}
\bibliography{egbib}

\begin{thebibliography}{10}\itemsep=-1pt

\bibitem{abdi2010principal}
Herv{\'e} Abdi and Lynne~J Williams.
\newblock Principal component analysis.
\newblock {\em Wiley interdisciplinary reviews: computational statistics},
  2(4):433--459, 2010.

\bibitem{bian2019depth}
Jia-Wang Bian, Zhichao Li, Naiyan Wang, Huangying Zhan, Chunhua Shen, Ming-Ming
  Cheng, and Ian Reid.
\newblock Unsupervised scale-consistent depth and ego-motion learning from
  monocular video.
\newblock In {\em NeurIPS}, 2019.

\bibitem{bian2020unsupervised}
Jia-Wang Bian, Huangying Zhan, Naiyan Wang, Tat-Jun Chin, Chunhua Shen, and Ian
  Reid.
\newblock Unsupervised depth learning in challenging indoor video: Weak
  rectification to rescue.
\newblock {\em arXiv}, 2020.

\bibitem{casser2019struct2depth}
Vincent Casser, Soeren Pirk, Reza Mahjourian, and Anelia Angelova.
\newblock Depth prediction without the sensors: Leveraging structure for
  unsupervised learning from monocular videos.
\newblock In {\em AAAI}, 2019.

\bibitem{concha2014using}
Alejo Concha and Javier Civera.
\newblock Using superpixels in monocular slam.
\newblock In {\em ICRA}, pages 365--372. IEEE, 2014.

\bibitem{concha2015dpptam}
Alejo Concha and Javier Civera.
\newblock Dpptam: Dense piecewise planar tracking and mapping from a monocular
  sequence.
\newblock In {\em IROS}, pages 5686--5693. IEEE, 2015.

\bibitem{dai2017scannet}
Angela Dai, Angel~X Chang, Manolis Savva, Maciej Halber, Thomas Funkhouser, and
  Matthias Nie{\ss}ner.
\newblock Scannet: Richly-annotated 3d reconstructions of indoor scenes.
\newblock In {\em CVPR}, volume~1, 2017.

\bibitem{deng2009imagenet}
Jia Deng, Wei Dong, Richard Socher, Li-Jia Li, Kai Li, and Li Fei-Fei.
\newblock Imagenet: A large-scale hierarchical image database.
\newblock In {\em CVPR}. Ieee, 2009.

\bibitem{eigen2015predicting}
David Eigen and Rob Fergus.
\newblock Predicting depth, surface normals and semantic labels with a common
  multi-scale convolutional architecture.
\newblock In {\em ICCV}, 2015.

\bibitem{eigen2014depth}
David Eigen, Christian Puhrsch, and Rob Fergus.
\newblock Depth map prediction from a single image using a multi-scale deep
  network.
\newblock In {\em NeurIPS}, 2014.

\bibitem{felzenszwalb2004efficient}
Pedro~F Felzenszwalb and Daniel~P Huttenlocher.
\newblock Efficient graph-based image segmentation.
\newblock {\em International journal of computer vision}, 59(2):167--181, 2004.

\bibitem{fischler1981random}
Martin~A Fischler and Robert~C Bolles.
\newblock Random sample consensus: a paradigm for model fitting with
  applications to image analysis and automated cartography.
\newblock {\em Communications of the ACM}, 24(6):381--395, 1981.

\bibitem{fransens2006hierarchical}
Jan Fransens and Frank Van~Reeth.
\newblock Hierarchical pca decomposition of point clouds.
\newblock In {\em Third International Symposium on 3D Data Processing,
  Visualization, and Transmission (3DPVT'06)}, pages 591--598. IEEE, 2006.

\bibitem{fu2018deep}
Huan Fu, Mingming Gong, Chaohui Wang, Kayhan Batmanghelich, and Dacheng Tao.
\newblock Deep ordinal regression network for monocular depth estimation.
\newblock In {\em CVPR}, 2018.

\bibitem{garg2016unsupervised}
Ravi Garg, Vijay~Kumar BG, Gustavo Carneiro, and Ian Reid.
\newblock Unsupervised cnn for single view depth estimation: Geometry to the
  rescue.
\newblock In {\em ECCV}, 2016.

\bibitem{geiger2013vision}
Andreas Geiger, Philip Lenz, Christoph Stiller, and Raquel Urtasun.
\newblock Vision meets robotics: The kitti dataset.
\newblock {\em IJRR}, 32(11):1231--1237, 2013.

\bibitem{godard2017unsupervised}
Cl{\'e}ment Godard, Oisin Mac~Aodha, and Gabriel~J Brostow.
\newblock Unsupervised monocular depth estimation with left-right consistency.
\newblock In {\em CVPR}, 2017.

\bibitem{godard2019digging}
Cl{\'e}ment Godard, Oisin Mac~Aodha, Michael Firman, and Gabriel~J Brostow.
\newblock Digging into self-supervised monocular depth estimation.
\newblock In {\em ICCV}, 2019.

\bibitem{gomez2019pl}
Ruben Gomez-Ojeda, Francisco-Angel Moreno, David Zuniga-No{\"e}l, Davide
  Scaramuzza, and Javier Gonzalez-Jimenez.
\newblock Pl-slam: A stereo slam system through the combination of points and
  line segments.
\newblock {\em IEEE Transactions on Robotics}, 35(3):734--746, 2019.

\bibitem{gosavi2012form}
Abhijit Gosavi and Elizabeth Cudney.
\newblock Form errors in precision metrology: a survey of measurement
  techniques.
\newblock {\em Quality Engineering}, 24(3):369--380, 2012.

\bibitem{hartley2003multiple}
Richard Hartley and Andrew Zisserman.
\newblock {\em Multiple view geometry in computer vision {(2.} ed.)}.
\newblock Cambridge University Press, 2006.

\bibitem{he2016deep}
Kaiming He, Xiangyu Zhang, Shaoqing Ren, and Jian Sun.
\newblock Deep residual learning for image recognition.
\newblock In {\em CVPR}, pages 770--778, 2016.

\bibitem{hoppe1992surface}
Hugues Hoppe, Tony DeRose, Tom Duchamp, John McDonald, and Werner Stuetzle.
\newblock Surface reconstruction from unorganized points.
\newblock In {\em Proceedings of the 19th annual conference on computer
  graphics and interactive techniques}, pages 71--78, 1992.

\bibitem{hu2021boosting}
Junjie Hu, Chenyou Fan, Hualie Jiang, Xiyue Guo, Yuan Gao, Xiangyong Lu, and
  Tin~Lun Lam.
\newblock Boosting light-weight depth estimation via knowledge distillation.
\newblock {\em arXiv}, 2021.

\bibitem{hu2019revisiting}
Junjie Hu, Mete Ozay, Yan Zhang, and Takayuki Okatani.
\newblock Revisiting single image depth estimation: Toward higher resolution
  maps with accurate object boundaries.
\newblock In {\em WACV}, 2019.

\bibitem{jaderberg2015spatial}
Max Jaderberg, Karen Simonyan, Andrew Zisserman, et~al.
\newblock Spatial transformer networks.
\newblock In {\em NeurIPS}, 2015.

\bibitem{jiang2020dipe}
Hualie Jiang, Laiyan Ding, Zhenglong Sun, and Rui Huang.
\newblock Dipe: Deeper into photometric errors for unsupervised learning of
  depth and ego-motion from monocular videos.
\newblock In {\em IROS}, 2020.

\bibitem{jiang2021unsupervised}
Hualie Jiang, Laiyan Ding, Zhenglong Sun, and Rui Huang.
\newblock Unsupervised monocular depth perception: Focusing on moving objects.
\newblock {\em IEEE Sensors Journal}, 2021.

\bibitem{khoshelham2011accuracy}
Kourosh Khoshelham.
\newblock Accuracy analysis of kinect depth data.
\newblock In {\em ISPRS workshop laser scanning}, volume~38, 2011.

\bibitem{kingma2014adam}
Diederik~P Kingma and Jimmy Ba.
\newblock Adam: A method for stochastic optimization.
\newblock {\em arXiv preprint arXiv:1412.6980}, 2014.

\bibitem{klingner2020self}
Marvin Klingner, Jan-Aike Term{\"o}hlen, Jonas Mikolajczyk, and Tim
  Fingscheidt.
\newblock Self-supervised monocular depth estimation: Solving the dynamic
  object problem by semantic guidance.
\newblock In {\em ECCV}, pages 582--600. Springer, 2020.

\bibitem{laina2016deeper}
Iro Laina, Christian Rupprecht, Vasileios Belagiannis, Federico Tombari, and
  Nassir Navab.
\newblock Deeper depth prediction with fully convolutional residual networks.
\newblock In {\em 3DV}, 2016.

\bibitem{lee2021learning}
Seokju Lee, Sunghoon Im, Stephen Lin, and In~So Kweon.
\newblock Learning monocular depth in dynamic scenes via instance-aware
  projection consistency.
\newblock In {\em AAAI}, 2021.

\bibitem{lee2019elaborate}
Sang~Jun Lee and Sung~Soo Hwang.
\newblock Elaborate monocular point and line slam with robust initialization.
\newblock In {\em ICCV}, pages 1121--1129, 2019.

\bibitem{liu2019planercnn}
Chen Liu, Kihwan Kim, Jinwei Gu, Yasutaka Furukawa, and Jan Kautz.
\newblock Planercnn: 3d plane detection and reconstruction from a single image.
\newblock In {\em CVPR}, pages 4450--4459, 2019.

\bibitem{liu2018planenet}
Chen Liu, Jimei Yang, Duygu Ceylan, Ersin Yumer, and Yasutaka Furukawa.
\newblock Planenet: Piece-wise planar reconstruction from a single rgb image.
\newblock In {\em CVPR}, pages 2579--2588, 2018.

\bibitem{liu2016learning}
Fayao Liu, Chunhua Shen, Guosheng Lin, and Ian~D Reid.
\newblock Learning depth from single monocular images using deep convolutional
  neural fields.
\newblock {\em IEEE TPAMI}, 38(10):2024--2039, 2016.

\bibitem{luo2019every}
Chenxu Luo, Zhenheng Yang, Peng Wang, Yang Wang, Wei Xu, Ram Nevatia, and Alan
  Yuille.
\newblock Every pixel counts++: Joint learning of geometry and motion with 3d
  holistic understanding.
\newblock {\em IEEE TPAMI}, 42(10):2624--2641, 2019.

\bibitem{meer1991robust}
Peter Meer, Doron Mintz, Azriel Rosenfeld, and Dong~Yoon Kim.
\newblock Robust regression methods for computer vision: A review.
\newblock {\em International journal of computer vision}, 6(1):59--70, 1991.

\bibitem{nievergelt1994total}
Yves Nievergelt.
\newblock Total least squares: State-of-the-art regression in numerical
  analysis.
\newblock {\em SIAM review}, 36(2):258--264, 1994.

\bibitem{pumarola2017pl}
Albert Pumarola, Alexander Vakhitov, Antonio Agudo, Alberto Sanfeliu, and
  Francese Moreno-Noguer.
\newblock Pl-slam: Real-time monocular visual slam with points and lines.
\newblock In {\em ICRA}, pages 4503--4508. IEEE, 2017.

\bibitem{saxena2009make3d}
Ashutosh Saxena, Min Sun, and Andrew~Y Ng.
\newblock Make3d: Learning 3d scene structure from a single still image.
\newblock {\em IEEE TPAMI}, 31(5):824--840, 2009.

\bibitem{silberman2012indoor}
Nathan Silberman, Derek Hoiem, Pushmeet Kohli, and Rob Fergus.
\newblock Indoor segmentation and support inference from rgbd images.
\newblock In {\em ECCV}. Springer, 2012.

\bibitem{teed2020deepv2d}
Zachary Teed and Jia Deng.
\newblock Deepv2d: Video to depth with differentiable structure from motion.
\newblock 2020.

\bibitem{von2008lsd}
Rafael~Grompone Von~Gioi, Jeremie Jakubowicz, Jean-Michel Morel, and Gregory
  Randall.
\newblock Lsd: A fast line segment detector with a false detection control.
\newblock {\em IEEE transactions on pattern analysis and machine intelligence},
  32(4):722--732, 2008.

\bibitem{von2012lsd}
Rafael~Grompone Von~Gioi, J{\'e}r{\'e}mie Jakubowicz, Jean-Michel Morel, and
  Gregory Randall.
\newblock Lsd: a line segment detector.
\newblock {\em Image Processing On Line}, 2:35--55, 2012.

\bibitem{wang2018learning}
Chaoyang Wang, Jos{\'e} Miguel~Buenaposada, Rui Zhu, and Simon Lucey.
\newblock Learning depth from monocular videos using direct methods.
\newblock In {\em CVPR}, 2018.

\bibitem{wang2020vplnet}
Rui Wang, David Geraghty, Kevin Matzen, Richard Szeliski, and Jan-Michael
  Frahm.
\newblock Vplnet: Deep single view normal estimation with vanishing points and
  lines.
\newblock In {\em CVPR}, pages 689--698, 2020.

\bibitem{wang2004image}
Zhou Wang, Alan~C Bovik, Hamid~R Sheikh, Eero~P Simoncelli, et~al.
\newblock Image quality assessment: from error visibility to structural
  similarity.
\newblock {\em IEEE TIP}, 13(4):600--612, 2004.

\bibitem{xu2017multi}
Dan Xu, Elisa Ricci, Wanli Ouyang, Xiaogang Wang, and Nicu Sebe.
\newblock Multi-scale continuous crfs as sequential deep networks for monocular
  depth estimation.
\newblock In {\em CVPR}, 2017.

\bibitem{yang2018recovering}
Fengting Yang and Zihan Zhou.
\newblock Recovering 3d planes from a single image via convolutional neural
  networks.
\newblock In {\em ECCV}, pages 85--100, 2018.

\bibitem{yin2019enforcing}
Wei Yin, Yifan Liu, Chunhua Shen, and Youliang Yan.
\newblock Enforcing geometric constraints of virtual normal for depth
  prediction.
\newblock In {\em ICCV}, 2019.

\bibitem{yin2018geonet}
Zhichao Yin and Jianping Shi.
\newblock Geonet: Unsupervised learning of dense depth, optical flow and camera
  pose.
\newblock In {\em CVPR}, 2018.

\bibitem{IndoorSfMLearner}
Zehao Yu, Lei Jin, and Shenghua Gao.
\newblock P$^{2}$net: Patch-match and plane-regularization for unsupervised
  indoor depth estimation.
\newblock In {\em ECCV}, 2020.

\bibitem{yu2019single}
Zehao Yu, Jia Zheng, Dongze Lian, Zihan Zhou, and Shenghua Gao.
\newblock Single-image piece-wise planar 3d reconstruction via associative
  embedding.
\newblock In {\em CVPR}, pages 1029--1037, 2019.

\bibitem{zhang2012microsoft}
Zhengyou Zhang.
\newblock Microsoft kinect sensor and its effect.
\newblock {\em IEEE multimedia}, 19(2):4--10, 2012.

\bibitem{zhao2020towards}
Wang Zhao, Shaohui Liu, Yezhi Shu, and Yong-Jin Liu.
\newblock Towards better generalization: Joint depth-pose learning without
  posenet.
\newblock In {\em CVPR}, 2020.

\bibitem{zhou2015structslam}
Huizhong Zhou, Danping Zou, Ling Pei, Rendong Ying, Peilin Liu, and Wenxian Yu.
\newblock Structslam: Visual slam with building structure lines.
\newblock {\em IEEE Transactions on Vehicular Technology}, 64(4):1364--1375,
  2015.

\bibitem{zhou2019moving}
Junsheng Zhou, Yuwang Wang, Kaihuai Qin, and Wenjun Zeng.
\newblock Moving indoor: Unsupervised video depth learning in challenging
  environments.
\newblock In {\em ICCV}, 2019.

\bibitem{zhou2018open3d}
Qian-Yi Zhou, Jaesik Park, and Vladlen Koltun.
\newblock Open3d: A modern library for 3d data processing.
\newblock {\em arXiv preprint arXiv:1801.09847}, 2018.

\bibitem{zhou2017unsupervised}
Tinghui Zhou, Matthew Brown, Noah Snavely, and David~G Lowe.
\newblock Unsupervised learning of depth and ego-motion from video.
\newblock In {\em CVPR}, 2017.

\end{thebibliography}
}

\clearpage
\begin{figure*}[htp]
\begin{center}
\includegraphics[width=\linewidth] {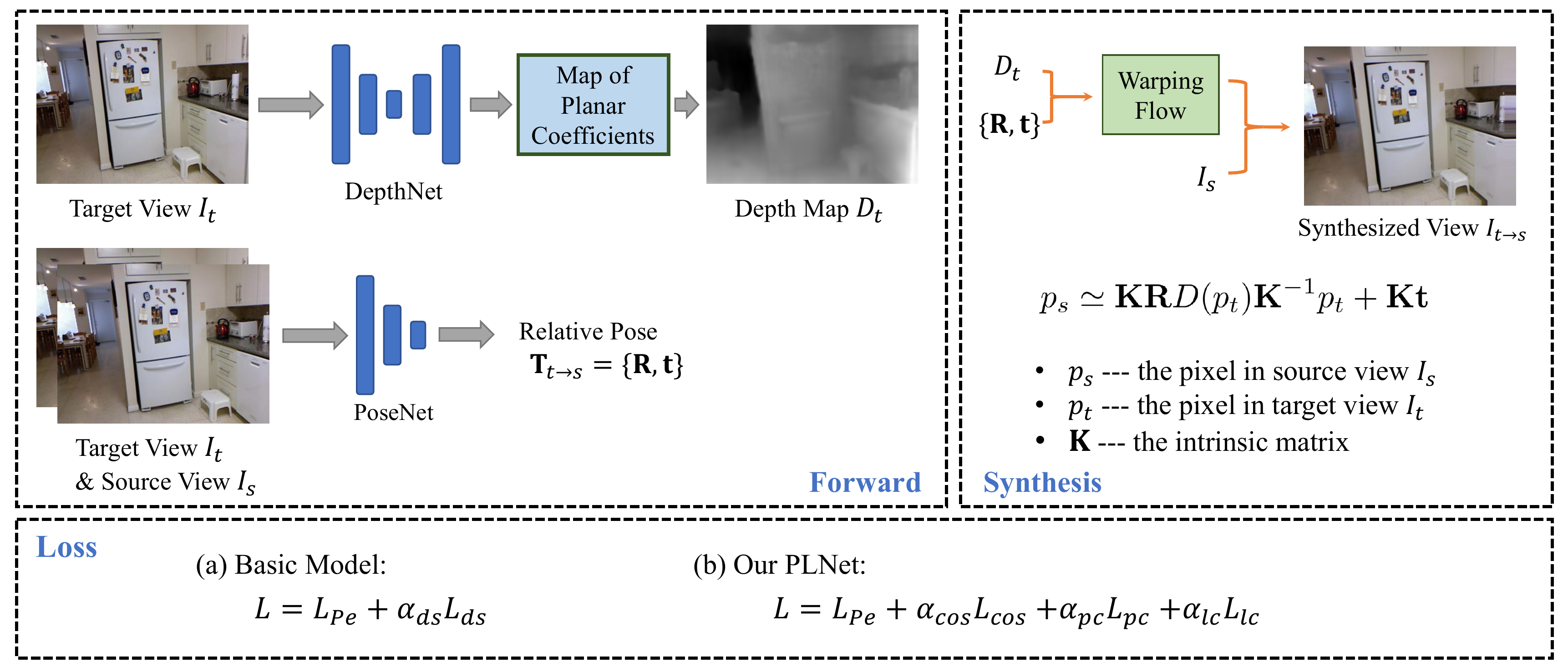}
\end{center}
\vspace*{-5mm}
\caption{\textbf{The Overall Framework of Unsupervised Monocular Depth Learning.}}
\label{fig:framework}
\vspace*{-2pt}
\end{figure*}

\begin{appendices}

\section{Overall Framework}
The overall framework of unsupervised monocular depth learning is illustrated in Fig.~\ref{fig:framework}. 
The framework contains three phases, network forward, view synthesis, and loss computation.
There are two networks. The first one is a DepthNet. Conventionally, it takes the target view $I_t$ as input and outputs the depth map $D_t$. But in our PLNet, DepthNet predicts the map of planar coefficients and converts it to the depth map with Eqn. 6 in the main paper. The second one is a Pose Net. Its inputs are the target view $I_t$ and the adjacent source view $I_s$. The output is the relative pose of the two views, ${\mathbf T}_{t \to s}$, including a rotation matrix $\mathbf R$ and a translation vector $\mathbf t$. After we have the depth map and relative pose, we can compute the warping flow from the target view to the source view. Next, we can synthesize the target view from the source view using the warping flow. Finally, we compute the photometric loss $L_{pe}$ between the target view and the synthesized view, and other regularization losses, $L_{ds}$, $L_{cos}$, $L_{pc}$ and $L_{lc}$, as the supervisory signal. Fig.~\ref{fig:framework} also  shows a comparison of traditional basic model and our proposed PLNet in the choice of losses.

\section{Empirical Analysis of Robustness}

This section gives empirical analysis above robustness of two strategies in enforcing plane regularization, \ie, the $L_{ssp}$ in P$^2$Net~\cite{IndoorSfMLearner} using least square fitting and the proposed $L_{pc}$ via random sample consistency. We use synthetic planer points with outliers. To be specific, we control the deviation of outliers to see how $L_{ssp}$ and $L_{pc}$ change. 

\begin{figure}[htp]
\centering
\begin{subfigure}{0.47\textwidth}
\includegraphics[width=0.98\linewidth]{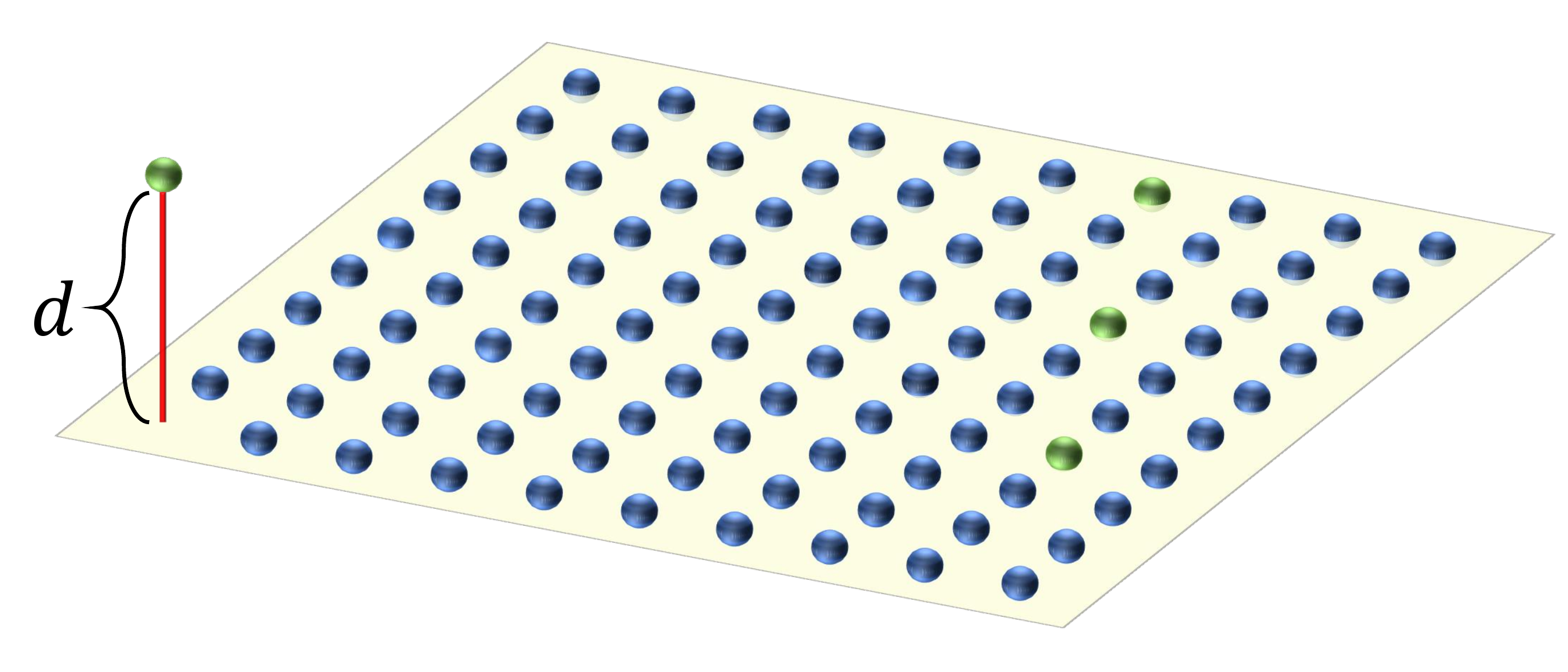}
\vspace{-3mm}
\caption{\small{(a) Visualization of grid points with a boundary outlier.}}
\end{subfigure}
\vspace{2mm}

\begin{subfigure}{0.5\textwidth}
\includegraphics[width=0.98\linewidth]{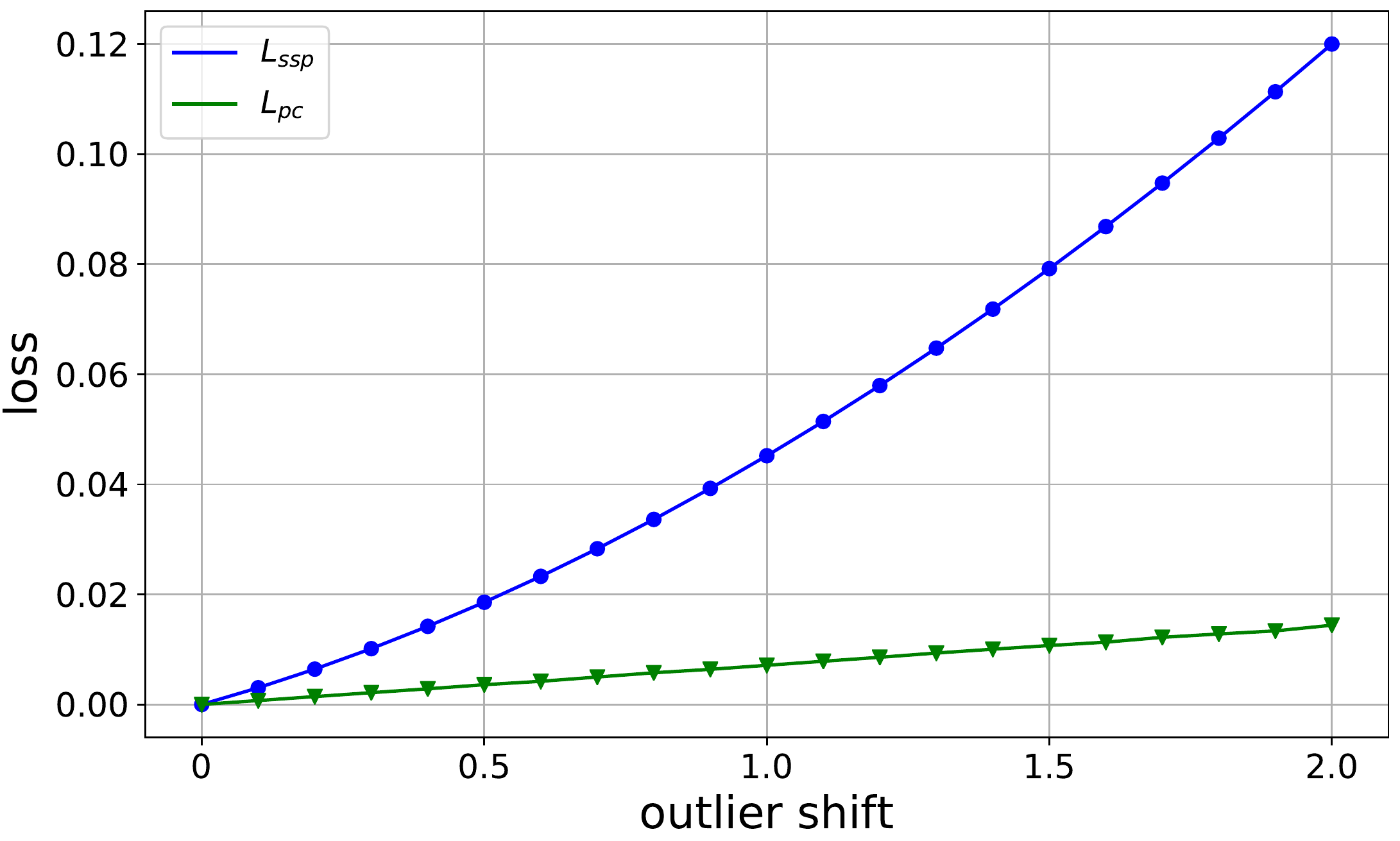}
\vspace*{-2mm}
\caption{\small{(b) The change of $L_{ssp}$ and $L_{pc}$ with the outlier shift}}
\end{subfigure}
\vspace{1mm}

\begin{subfigure}{0.5\textwidth}
\includegraphics[width=0.98\linewidth]{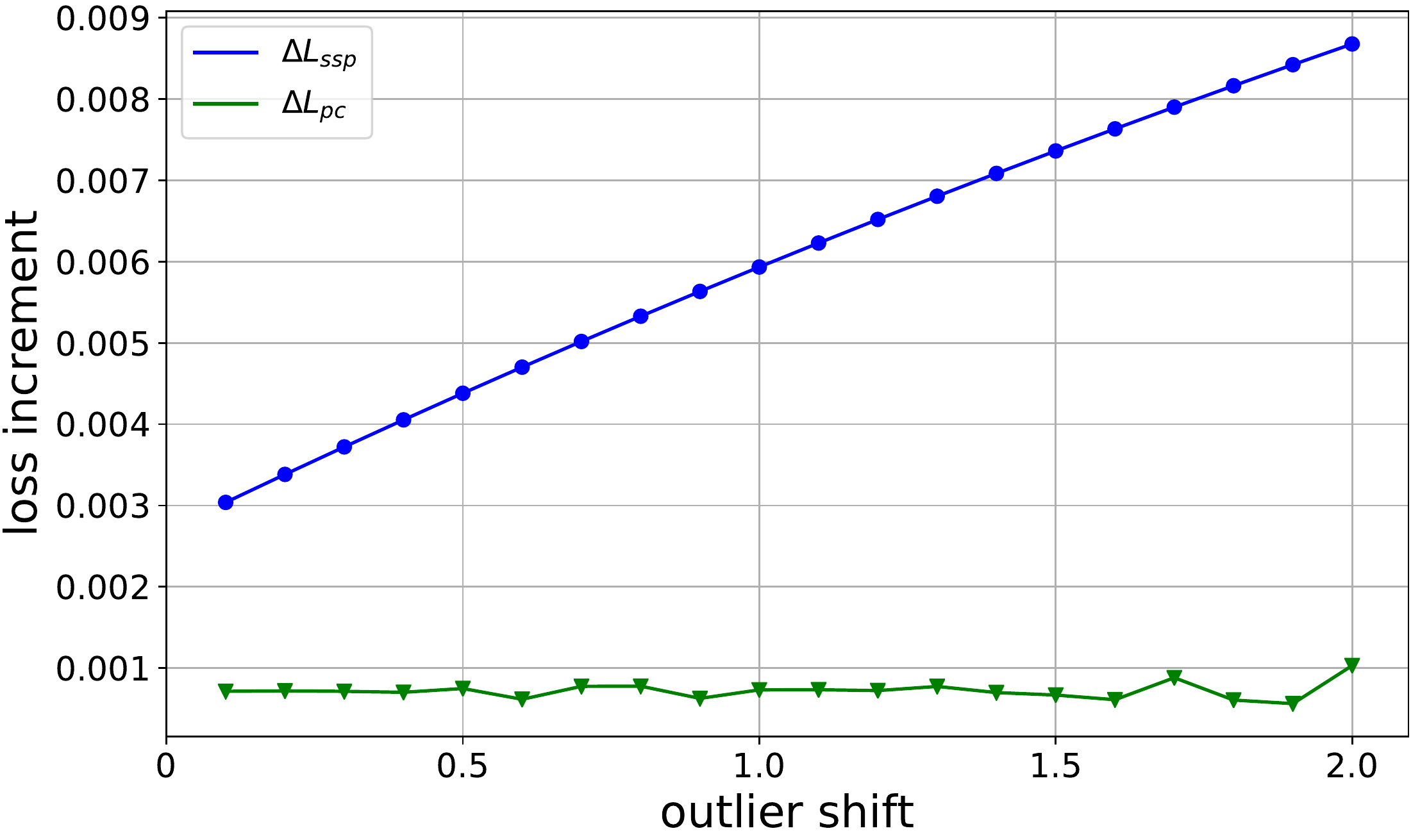}
\vspace*{-2mm}
\caption{\small{(c) The change of $\Delta L_{ssp}$ and $\Delta L_{pc}$ with the outlier shift}} 
\end{subfigure}
\caption{\textbf{Planar Points with A Boundary Outlier.}} \label{fig:boundary}
\end{figure}

\begin{figure}[htp]
\centering
\begin{subfigure}{0.47\textwidth}
\includegraphics[width=0.98\linewidth]{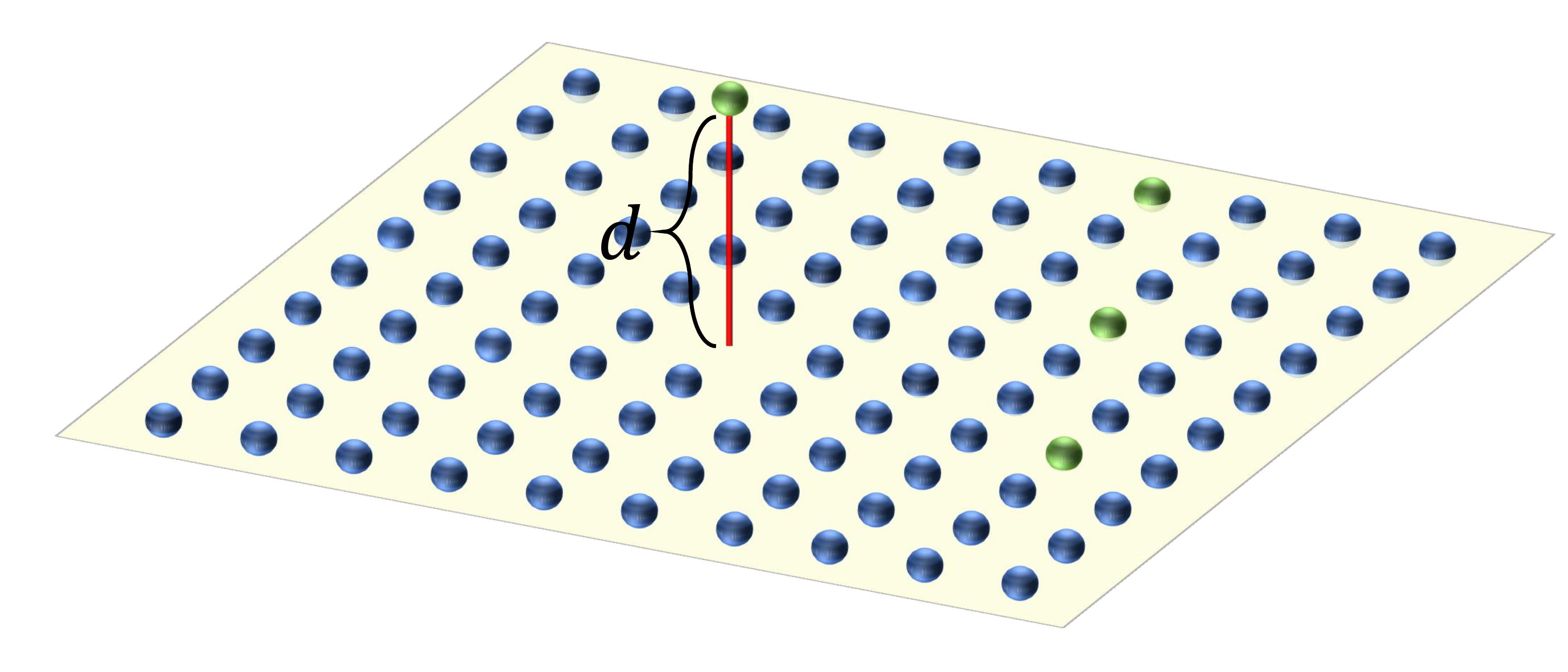}
\vspace{-3mm}
\caption{\small{(a) Visualization of grid points with a middle outlier.}} 
\end{subfigure}
\vspace{2mm}

\begin{subfigure}{0.5\textwidth}
\includegraphics[width=0.98\linewidth]{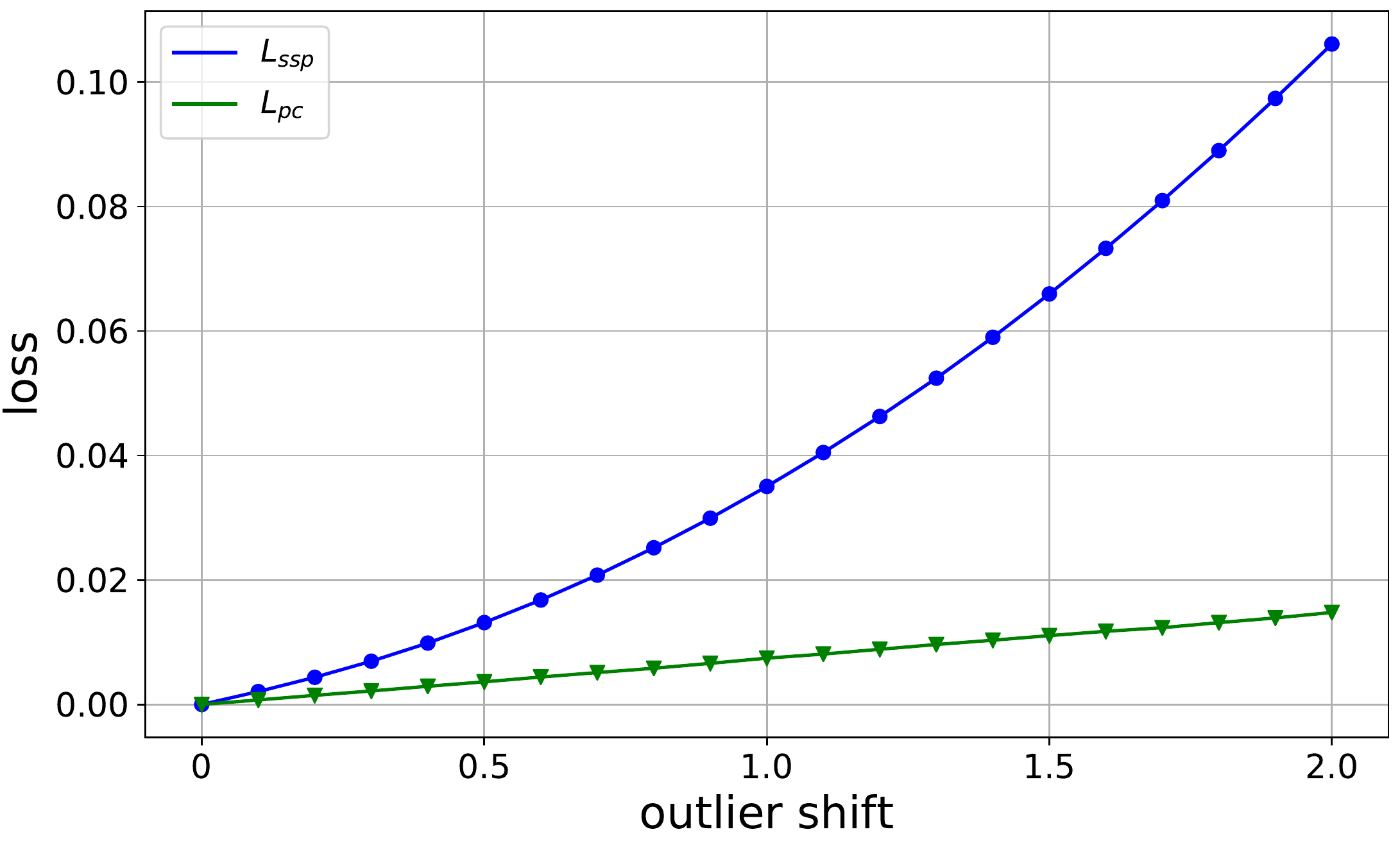}
\vspace*{-2mm}
\caption{\small{(b) The change of $L_{ssp}$ and $L_{pc}$ with the outlier shift}}
\end{subfigure}
\vspace{1mm}

\begin{subfigure}{0.5\textwidth}
\includegraphics[width=0.98\linewidth]{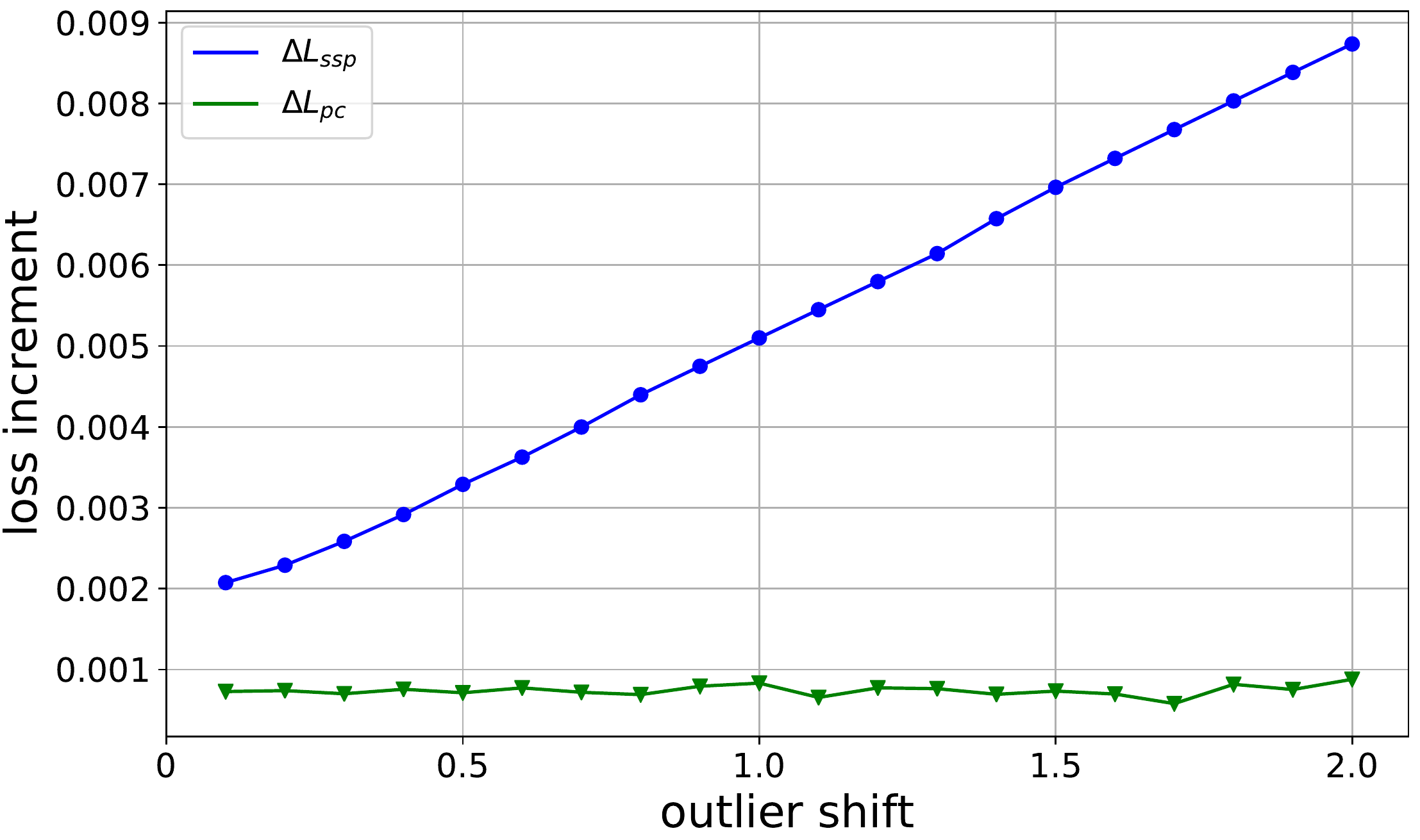}
\vspace*{-2mm}
\caption{\small{(c) The change of $\Delta L_{ssp}$ and $\Delta L_{pc}$ with the outlier shift}} 
\end{subfigure}
\caption{\textbf{Planar Points with A Middle Outlier.}} \label{fig:center}
\end{figure}

The synthesized data is a $10\times10$ grid of points parallel to the X-Y plane, as shown in (a) of Fig.~\ref{fig:boundary} and Fig.~\ref{fig:center}. 
The size of the grid is $1\times1$. 
Considering that the superpixels segmentation algorithm~\cite{felzenszwalb2004efficient} could produce the plane outlier in the region boundaries, we first simulate the outlier by shifting a point at the corner as shown in (a) of Fig.~\ref{fig:boundary}. 
For more complete verification, we also experiment with an outlier in the middle of the grid, as (a) of Fig.~\ref{fig:center}. 
We shift the side points by every 0.1 from 0 to 2. The larger the shift is, the more deviated the outliers are. To ensure the obtained $L_{pc}$ approximate to its expectation, we sample $10^6$ 4-points sets in computation.

How $L_{ssp}$ and $L_{pc}$ change with the outlier shift is shown in (b) of Fig.~\ref{fig:boundary} and Fig.~\ref{fig:center}. It can be observed that $L_{ssp}$ increases much faster than our $L_{pc}$, which indicates least square fitting is more sensitive to the outlier than the proposed random sample consistency strategy. To better illustrate how losses quantitatively increase with the outlier shift, we also plot the loss increment, $\Delta L(d_i) = L(d_i)- L(d_{i-1})$, in (c) of Fig.~\ref{fig:boundary} and Fig.~\ref{fig:center}. $\Delta L_{pc}$ is approximately constant under 0.001, indicating that $L_{pc}$ linearly increases with the outlier shift. In contrast, $\Delta L_{ssp}$ grows nearly linearly with the outlier shift, showing that $L_{ssp}$ increases quadratically. Therefore, $L_{pc}$ is less sensitive to the outlier than $L_{ssp}$. 

The planar parameters by least-square fitting could be severely biased by the outlier, causing $L_{ssp}$ to increase fast. As the outlier is the minority, it has a high probability that the sampled 4 points are from the inliers. Further, when the 4-points set contains the outlier, such as the \textcolor{green}{green} points in (a) of Fig.~\ref{fig:boundary} and Fig.~\ref{fig:center}, it is still possible that they are coplanar. That's the reason why $L_{pc}$ grows much slower.

\section{Depth Accuracy and Structure Regularity}

\begin{figure*}[h]
\centering
\begin{subfigure}{0.49\textwidth}
\includegraphics[width=0.96\linewidth]{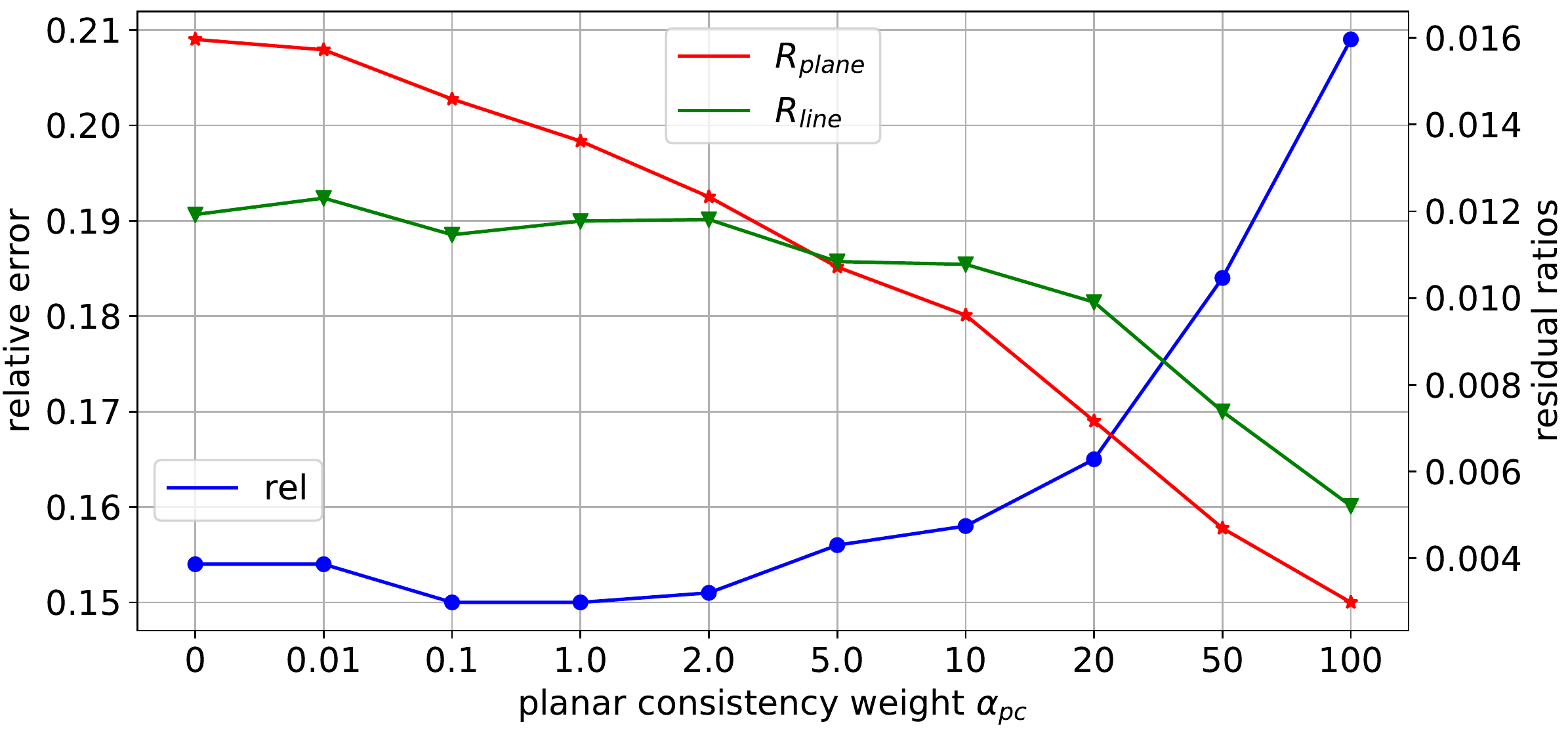}
\vspace*{-2mm}
\caption{\small{(a) The change of rel, ${R_{plane}}$ and ${R_{line}}$ with $\alpha_{pc}$,}} \label{fig:alpha_plane}
\vspace{-5pt}
\caption{\small{when $\alpha_{cos}$ and $\alpha_{lc}$ are fixed as 0.2 and 0.5.}}
\end{subfigure}
\hspace{-2mm}
\begin{subfigure}{0.49\textwidth}
\includegraphics[width=0.96\linewidth]{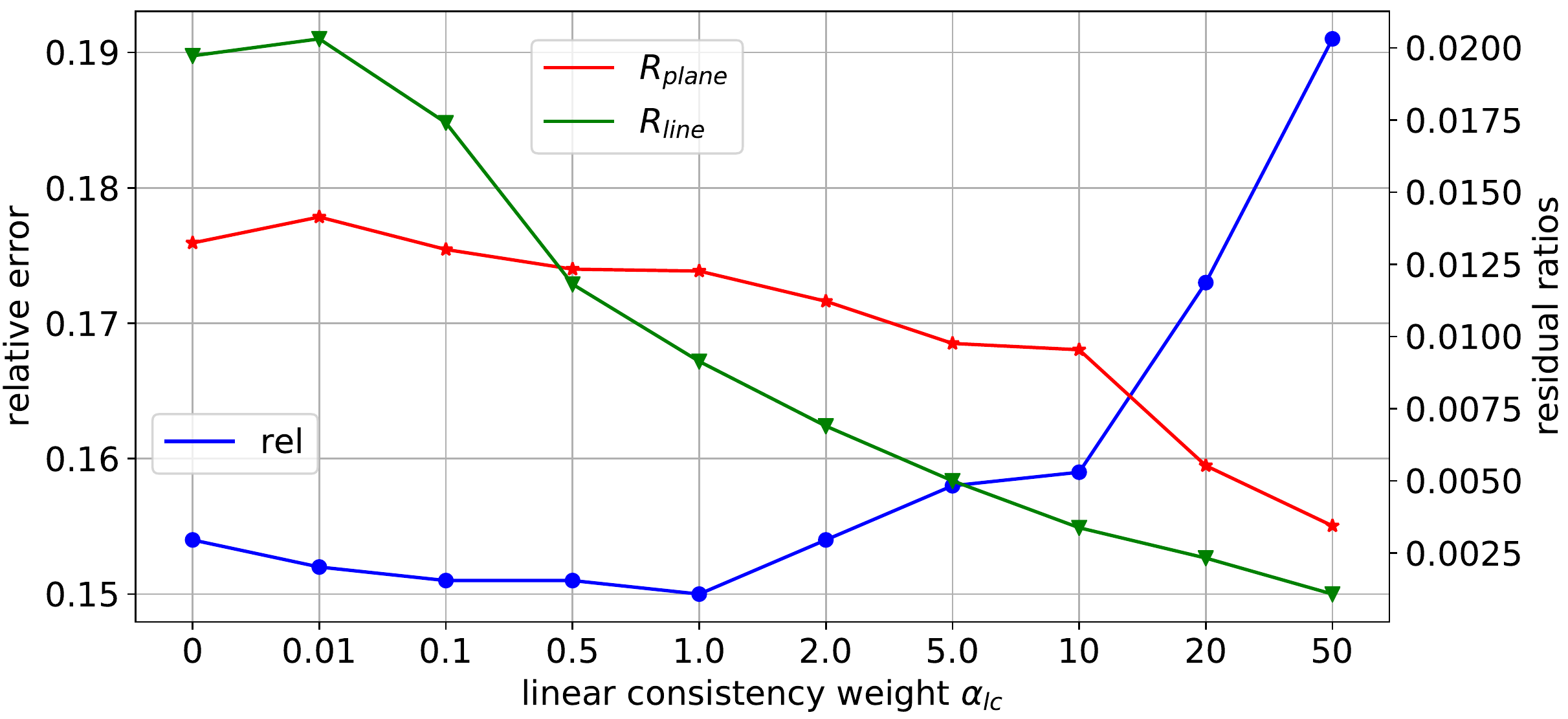}
\vspace*{-2mm}
\caption{\small{(b) The change of rel, ${R_{plane}}$ and ${R_{line}}$ with $\alpha_{lc}$,}} \label{fig:alpha_line}
\vspace{-5pt}
\caption{\small{when $\alpha_{cos}$ and $\alpha_{pc}$ are fixed as 0.2 and 2.0.}}
\end{subfigure}
\vspace*{-2mm}
\caption{\textbf{The Change of Depth Accuracy and Structure Regularity with the Consistency Weights.} } \label{fig:change}
\end{figure*}

This section shows how the depth accuracy and structure regularity change with the weights of the consistency loss functions. The results are shown in Fig.~\ref{fig:change}. For (a) of Fig.~\ref{fig:change}, we fix the weights of $L_{cos}$ and $L_{lc}$ as 0.2 and 0.5, and change the weight of $L_{pc}$. For (b) of Fig.~\ref{fig:change}, we fix the weights of $L_{cos}$ and $L_{pc}$ as 0.2 and 2.0, and change the weight of $L_{lc}$. 

When increasing either of the consistency weights, the flatness residual ratio $R_{plane}$ and straightness residual ratio $R_{line}$ tend to decrease. The planar consistency weight $\alpha_{pc}$ is more effective to the flatness, and the linear consistency weight $\alpha_{pc}$ is more effective to the straightness. The relative error of depth initially decreases when both consistency weights are smaller than 1.0. Afterward, the relative error begins to increase, but it is under 0.16 if the weights are not bigger than 10.  It is observed that when the consistency weights are too large, the predicted depth map is over smooth, and the resulted point cloud is over flattened. We can better trade-off between depth accuracy and structure regularity by finding a weight from 0.1 to 10 for the consistency loss functions.

\begin{table}[htp]
 \centering
 \resizebox{0.95\columnwidth}{!}{
    \begin{tabular}{c|c| ccc}
        \toprule[1pt]
        Methods & Supervised  & rot(deg) $\downarrow$& tr(deg) $\downarrow$& tr(cm) $\downarrow$\\ \hline
        DeepV2D~\cite{teed2020deepv2d} & \Checkmark & 0.714 & 12.205 &  1.514 \\
        \hline
        Monodepth2~\cite{godard2019digging} & \XSolidBrush &   1.914  &  36.878  &   4.916  \\
        P$^2$Net (\textit{3 frames})~\cite{IndoorSfMLearner} & \XSolidBrush &   1.850  &  35.242  &   4.484  \\
        P$^2$Net (\textit{5 frames})~\cite{IndoorSfMLearner} & \XSolidBrush &   1.860  &  35.112  &   4.433  \\
        \hline
        PLNet (\textit{3 frames}) & \XSolidBrush & 1.789  &  33.453  &   4.304  \\
        PLNet (\textit{5 frames}) & \XSolidBrush & \textbf{1.767}  &  \textbf{32.731}  & \textbf{4.220}  \\
       \toprule[1pt]
    \end{tabular}
 }
\caption{\textbf{Camera Pose Estimation Results.}}
\label{tab:pose}
\end{table}

\section{Pose Estimation on ScanNet}

We follow DeepV2D~\cite{teed2020deepv2d} and P$^2$Net~\cite{IndoorSfMLearner} to evaluate pose estimation on ScanNet~\cite{dai2017scannet} with the pre-trained model on NYU Depth V2~\cite{silberman2012indoor}. The results are listed in Tab.~\ref{tab:pose}. PLNet also clearly outperforms other unsupervised methods.

\section{Visualization of ScanNet Depth Maps}
\begin{figure*}[t]
\centering{
\input{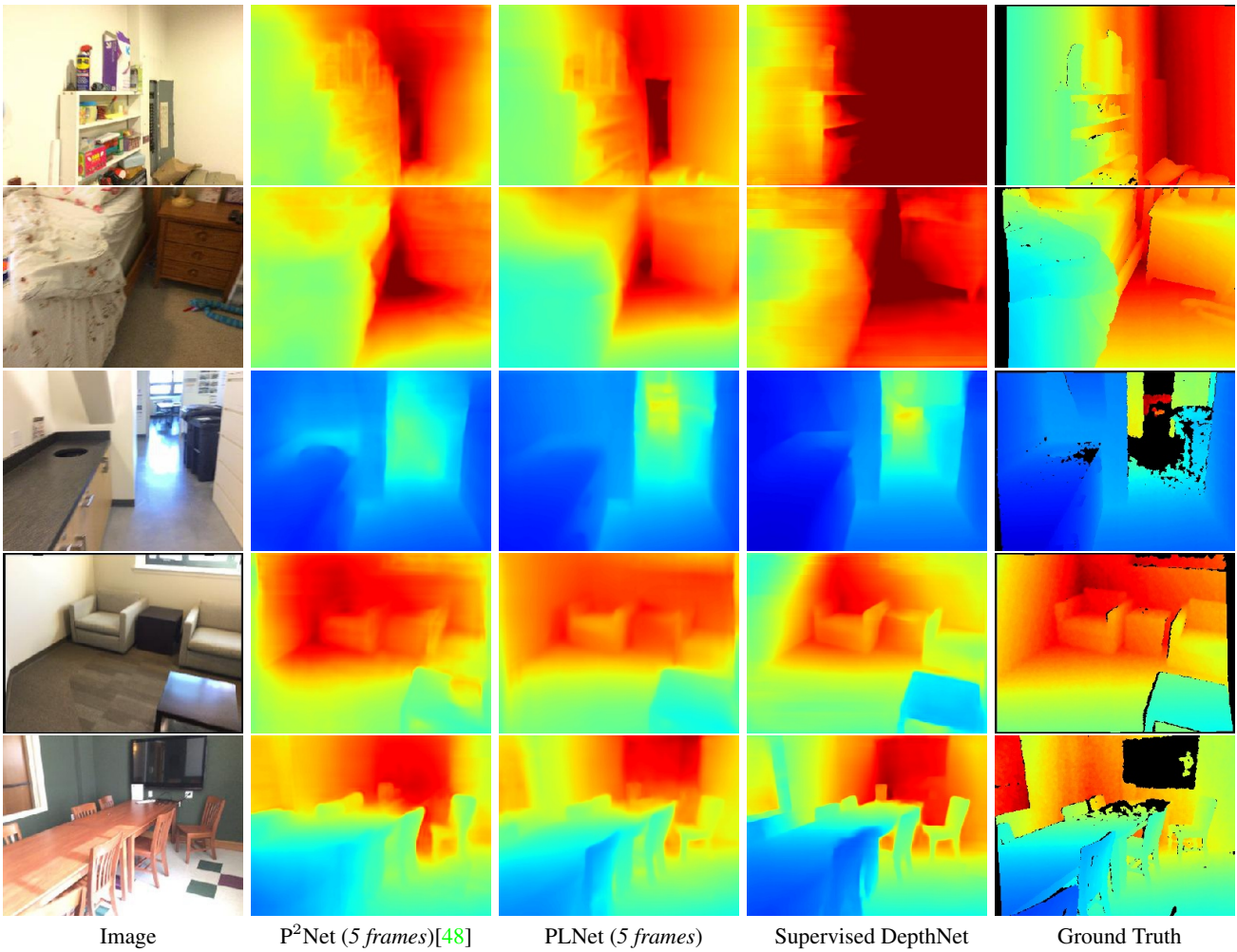}
}
\vspace{-2.5pt}
\caption{{\bf Predicted Depth Maps of ScanNet~\cite{dai2017scannet}.} Best viewed when zooming in.}
\label{fig:scannet_depths}
\end{figure*}

We visualize the depth maps of ScanNet~\cite{dai2017scannet} in Fig.~\ref{fig:scannet_depths}, which corresponds to the experiments on Sec.~5.5 of the manuscript. We do not perform alignment for the supervised DepthNet. The predicted depth maps are normalized by the maximum depth of the corresponding ground truth depth map. It is clear that for the first two examples, the supervised DepthNet overestimates the scene scale.

\end{appendices}

\end{document}